\documentclass[journal]{IEEEtran}
\usepackage{array}
\usepackage{times}
\usepackage{epsfig}
\usepackage{graphicx}
\usepackage{amsmath}
\usepackage{multirow}
\usepackage{amssymb}
\usepackage{enumitem}
\usepackage{booktabs}
\usepackage{longtable}
\usepackage{xcolor}
\usepackage{hyperref}
\ifCLASSINFOpdf
\else
\fi
\newcommand{\etal}{\textit{et al}. }

\begin{document}
%
\title{Self-supervised Learning of Detailed 3D Face Reconstruction}
%
%
%

\author{Yajing~Chen,
        Fanzi~Wu,
        Zeyu~Wang,
        Yibing~Song,
        Yonggen~Ling,
        and~Linchao~Bao$^*$
\thanks{Y. Chen, Y. Song, Y. Ling and L. Bao are with Tencent AI Lab, Shenzhen 518057, China (e-mail: jadechancyj907@gmail.com, yibingsong.cv@gmail.com, ylingaa@connect.ust.hk, linchaobao@gmail.com).}
\thanks{F. Wu is with Department of Electronic Engineering, The Chinese University of Hong Kong (e-mail: fzwu@link.cuhk.edu.hk).}
\thanks{Z. Wang is with Department of Computer Science, Columbia University (e-mail: zw2723@columbia.edu).}
\thanks{*L. Bao is the corresponding author.}}
\maketitle

\begin{abstract}
In this paper, we present an end-to-end learning framework for detailed 3D face reconstruction from a single image\footnote{Code is available at: \url{ https://github.com/cyj907/unsupervised-detail-layer}}.
Our approach uses a 3DMM-based coarse model and a displacement map in UV-space to represent a 3D face.
Unlike previous work addressing the problem, our learning framework does not require supervision of surrogate ground-truth 3D models computed with traditional approaches.
Instead, we utilize the input image itself as supervision during learning.
In the first stage, we combine a photometric loss and a facial perceptual loss between the input face and the rendered face, to regress a 3DMM-based coarse model.
In the second stage, both the input image and the regressed texture of the coarse model are unwrapped into UV-space, and then sent through an image-to-image translation network to predict a displacement map in UV-space. The displacement map and the coarse model are used to render a final detailed face, which again can be compared with the original input image to serve as a photometric loss for the second stage.
The advantage of learning displacement map in UV-space is that face alignment can be explicitly done during the unwrapping, thus facial details are easier to learn from large amount of data. Extensive experiments demonstrate the superiority of our method over previous work.
\end{abstract}


%
\IEEEpeerreviewmaketitle

\section{Introduction}
%
%
%
%

\IEEEPARstart{R}{ecovering} the 3D human facial geometry from a single color image is an ill-posed problem. Existing methods typically employ a parametric face modeling framework named as 3D morphable model (3DMM)  \cite{blanz1999morphable}. In a 3DMM there are a set of facial shapes and texture bases, which are built from real-world 3D face scans. A linear combination of these bases synthesizes a 3D face model. During the training process, a loss function is constructed to measure the difference between the input face image and the 3D face models. The linear coefficients (i.e., 3DMM parameters) can be generated by minimizing the computed loss. While conventional methods learn these coefficients via analysis-by-synthesis optimization  \cite{blanz2003face,romdhani2005estimating}, recent studies have shown the effectiveness of regressing 3DMM parameters using CNN based approaches  \cite{zhu2016face,tran2017regressing,tewari2017mofa,kim2018inversefacenet,genova2018unsupervised}.

\begin{figure}[t]
   \includegraphics[width=\linewidth]{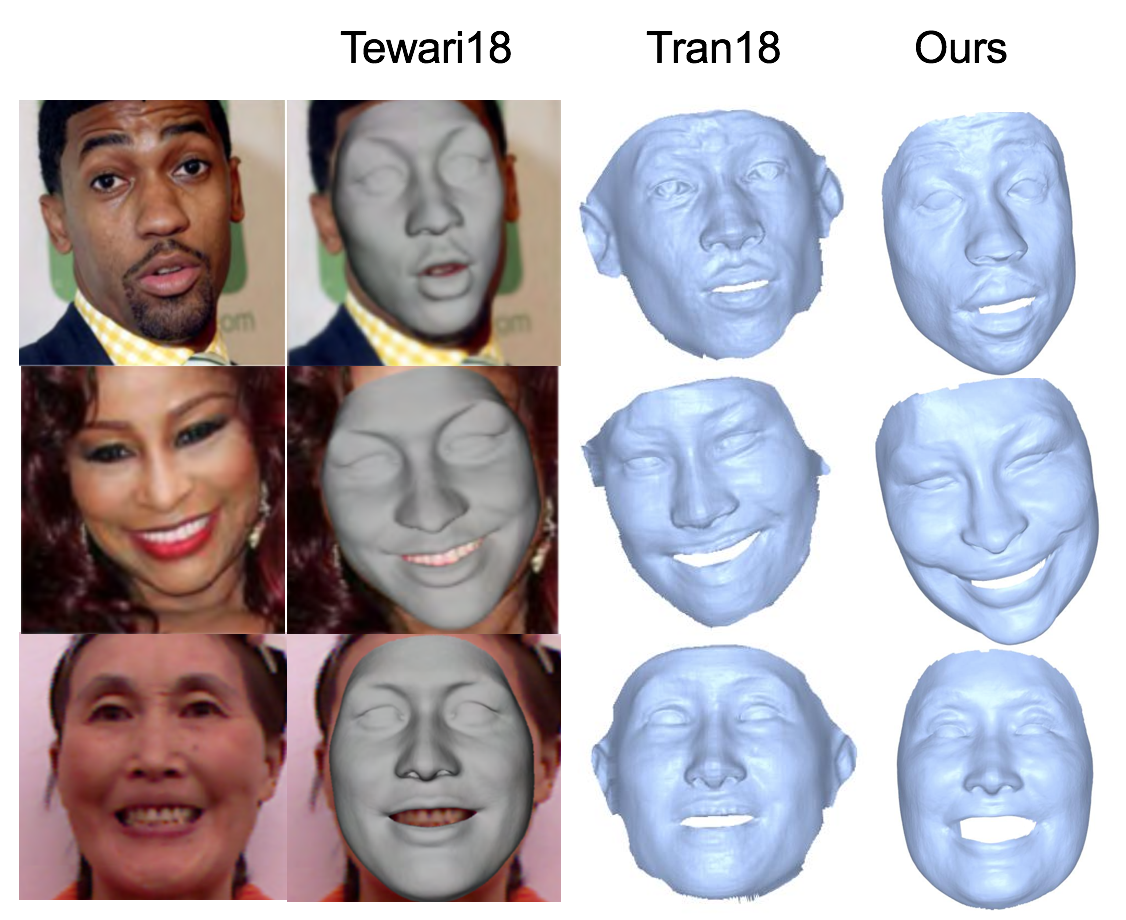}
    \caption{Our method can produce more faithful 3D face models than state-of-the-art methods like Tewari18  \cite{Tewari_2018_CVPR} and Tran18  \cite{tran2018extreme}.}
    \label{fig:teaser}
\end{figure}

Learning to regress 3DMM parameters via CNN requires a large amount of data. For methods based on supervised learning, the ground-truth 3DMM parameters are generated by optimization-based fitting  \cite{zhu2016face,tran2017regressing} or synthetic data generation  \cite{richardson20163d,dou2017end}. The limitations appear that the generated ground-truth labels are not accurate and the synthetic data lacks realism. In comparison, the methods  \cite{tewari2017mofa,genova2018unsupervised} based on self-supervised learning\footnote{We in this paper do not distinguish between the term ``self-supervised'' and ``unsupervised'', as both refer to learning without ground-truth annotations in our case. We prefer the term ``self-supervised''.} do not employ this process but learn directly from unlabeled face images. For example, MoFA  \cite{tewari2017mofa} learns to regress 3DMM parameters by forcing the rendered images to have similar pixel colors as input images in facial regions. However, enforcing the pixel level similarity does not imply similar facial identities.
Genova \etal  \cite{genova2018unsupervised} rendered images of a face from multiple views. They use a face recognition network to measure the perceptual similarity between the input faces and the rendered faces. 
Although the method is capable of producing 3D models resembling the faces in the input images, it ignores detailed facial characteristics and leads to unfaithful reconstructions. 

In order to model facial details beyond 3DMM, a few deep learning methods have been proposed recently.
For the methods  \cite{jackson2017large,sela2017unrestricted} represent 3D faces completely without using 3DMM, severely degraded results are usually obtained. More robust approaches typically represent 3D faces with detail modeling in addition to 3DMM  \cite{richardson2017learning,Tewari_2018_CVPR,tran2018extreme}.
For example, learned parametric correctives are employed in  \cite{Tewari_2018_CVPR}, and 3D detail maps are employed in  \cite{richardson2017learning,tran2018extreme}. Since the learned parametric correctives  \cite{Tewari_2018_CVPR} have very limited expressive capabilities (see Fig. \ref{fig:teaser}), we advocate 3D detail maps for detail modeling.
However, existing approaches employing detail maps  \cite{richardson2017learning,tran2018extreme} rely on surrogate ground-truth detail maps computed from traditional approaches, which are error prone and limit the fidelity of the reconstruction.

In this paper, we propose a two-stage framework to regress 3DMM parameters and reconstruct facial details via self-supervised learning. In the first stage, we use a combination of multi-level loss terms to train the 3DMM regression network. These loss terms consist of low-level photometric loss, mid-level facial landmark loss, and high-level facial perceptual loss, which enable the network to preserve both facial appearances and identities.
In the second stage, we employ an image-to-image translation network to capture the missing details. We unwrap both the input image and the regressed 3DMM texture maps into UV-space. The corresponding UV maps are together sent into the translation network to obtain the detailed displacement map in UV-space.
The displacement map and the 3DMM coarse model together are rendered to a final face image, which is enforced to be photometric consistent with the input face image during training.
Finally, the whole network can be trained end-to-end without any annotated labels.
%
The advantage of the detail modeling in UV-space is that all the training face images with different poses are aligned in UV-space, which facilitates the network to capture invariant details in spatial regions around facial components with large amount of data.
The main contribution of our work is that we use a self-supervised approach to solve a challenging task of detailed 3D face reconstruction from a single RGB image and we achieve very high quality results.
We conduct extensive experiments and analysis to show the effectiveness of our method.
Compared with state-of-the-art approaches, the 3D face models produced by our method are generally more faithful to the input face images.

\section{Related Work}

In this section, we briefly perform a literature survey on single-view 3D face reconstruction methods. These methods can be categorized as the optimization based, the supervised learning and the self-supervised learning based methods. A more complete review can be found in  \cite{zollhofer2018state}.

{\flushleft \bf 3DMM by Optimization.}
The 3D morphable model (3DMM) is proposed in  \cite{blanz1999morphable} to reconstruct a 3D face by a linear combination of shape and texture blendshapes. These blendshapes (i.e., bases) are extracted by PCA on aligned 3D face scans. Later, Cao \etal  \cite{cao2014facewarehouse} bring facial expressions into 3DMM and introduce a bilinear face model named FaceWarehouse. Since then, reconstructing a 3D face from an input image can be formulated as generating the optimal 3DMM parameters including shape, expression, and texture coefficients, such that the model-induced image is similar to the input image in the predefined feature spaces. Under this formulation, the analysis-by-synthesis optimization framework  \cite{blanz2003face,romdhani2005estimating,garrido2013reconstructing,thies2016face2face} is commonly adopted.
However, these optimization-based approaches are parameter sensitive. Non-realistic appearances exist on the generated 3D model.

{\flushleft \bf 3DMM by Supervised Learning.}
Methods based on supervised learning requires ground-truth 3DMM labels by either optimization-based fitting or synthetic data rendering. Zhu \etal  \cite{zhu2016face} and Tran \etal  \cite{tran2017regressing} use 3DMM parameters generated by optimization-based approaches as ground-truth to learn their CNN models. The performance of these methods are limited by unreliable labels. On the other hand, other approaches  \cite{dou2017end,richardson20163d,kim2018inversefacenet} try to utilize synthetic data rendered with random 3DMM parameters for supervised learning. Dou \etal  \cite{dou2017end} propose to use synthetic face images and corresponding 3D scans together for network learning. Richardson \etal  \cite{richardson20163d} train a 3DMM regression network with only synthetic rendered face images. Kim \etal  \cite{kim2018inversefacenet} show that training with synthetic data can be adapted to real data with the bootstrapping algorithm. However, the performances of these methods are limited by the unrealistic input and the 3D face models do not resemble the input images.

{\flushleft \bf 3DMM by Self-supervised Learning.}
Self-supervised methods derive supervisions by using input images without labels. MoFA  \cite{tewari2017mofa} uses a pixel-wise photometric loss to ensure the rendered image induced by the estimated 3DMM parameters to be similar to the input image. However, the photometric loss makes the network attend to pixel-wise similarity between rendered image and input image, while the identity of the input face is ignored. Recently, Genova \etal  \cite{genova2018unsupervised} propose to enforce the feature similarity between the rendered image and the input image via a fixed-weight face recognition network. Their model preserves facial identity, however the low-level features are not similar (e.g., illumination, skin color, and facial expressions). This is because it is designed to predict 3DMM parameters from illumination and expression invariant facial feature embeddings instead of original face images.
In the case of multi-view 3DMM regression, MVF-Net \cite{wu2019mvf} learns a deep network using multi-view face images with self-supervised view alignment loss.

{\flushleft \bf Detail Modeling beyond 3DMM.}
Due to the limited expressive power of 3DMM, some approaches try to model facial details with additional layers built upon 3DMM.
Examples following such type of modeling include the depth maps  \cite{richardson2017learning} or bump maps  \cite{tran2018extreme}, and trainable corrective models  \cite{Tewari_2018_CVPR}.
Besides, some other work employs non-parametric 3D representations  \cite{jackson2017large,sela2017unrestricted} to gain more degrees of freedoms, but usually are less robust than the methods built upon 3DMM.

In this paper, we focus on 3D representations with a coarse 3DMM model and an additional detail-layer. In the coarse model, different from MoFA  \cite{tewari2017mofa} and Genova \etal  \cite{genova2018unsupervised}, we combine the use of low-level photometric loss and high-level perceptual loss to provide multi-level supervision for the coarse model. For the detail-layer representation, we prefer detail maps rather than trainable corrective models  \cite{Tewari_2018_CVPR} due to more expressive power.
Note that existing detail map based approaches  \cite{richardson2017learning,tran2018extreme} are all based on supervised learning with surrogate ground-truth detail maps computed with traditional methods, while our approach is completely unsupervised. We use the unwrapped input image and 3D model in the UV space as inputs to the neural network, and make the network learn the detailed information from the difference between real and rendered images in an aligned space.
Different from Guo \etal  \cite{guo2019cnn} that used RGBD data as model inputs and learns the per-vertex normal displacement with UV maps, we use RGB images as inputs and build a UV render layer that builds dense correspondence between UV space and the rendered image space. These differences are vital for better face reconstruction.

\begin{figure*}[t]
    \centering
    \includegraphics[width=0.95\linewidth]{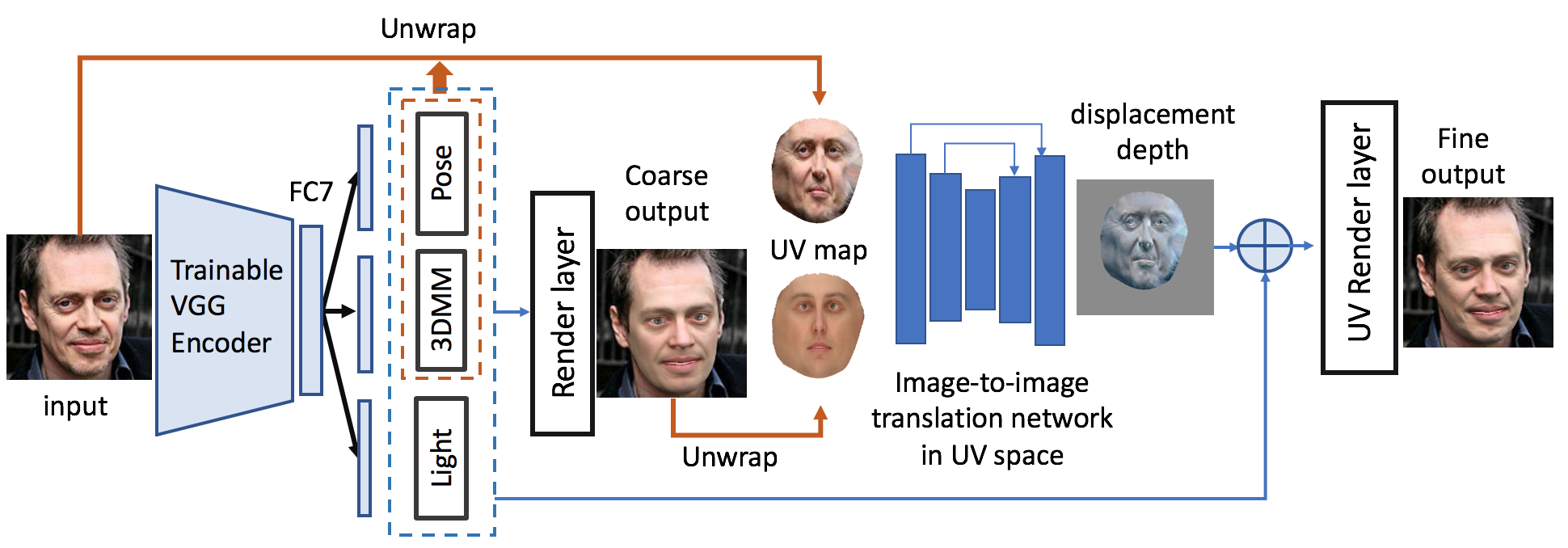}
    \caption{Proposed pipeline. We use a 3DMM encoder to transform an input face image into a latent code vector to regress the 3DMM parameters. We unwrap both the input image and the reconstructed 3D model into UV space and estimate a displacement depth map. Then, the 3DMM-based coarse model and the displacement depth map are used to generate a 3D face model with fine details.}
    \label{fig:pipeline}
\end{figure*}

\section{Proposed Method}

In this section, we first give an overview of our method, and then explain the details of each module.

\subsection{Framework Overview}

Figure \ref{fig:pipeline} illustrates the pipeline of the proposed framework. It consists of two modules. The first one is the 3DMM regression module and the second one is the detail modeling module. The 3DMM regression module learns to predict the 3DMM parameters from an input image with a trainable encoder network and a non-trainable differentiable renderer, which is similar to MoFA  \cite{tewari2017mofa}. 
The detail modeling module employs an image-to-image translation network to predict a displacement depth map in UV space from the unwrapped input image and the regressed 3DMM texture UV map. 
The displacement depth map is then added back to the regressed 3DMM-based coarse model to get the final detailed 3D model.
Finally, a differentiable UV renderer enables the whole learning process to be self-supervised by comparing the differences between the final rendered output and the input image. 

The training process consists of two stages. 
In the first stage, we train the 3DMM regression module without the detail modeling module using two types of self-supervised losses including pixel-level photometric loss \cite{tewari2017mofa} and perceptual-level identity loss \cite{genova2018unsupervised}. 
In the second stage, we fix the weights in the 3DMM regression module and train the detail modeling module using the pixel-level photometric loss, as well as additional smoothness losses and regularization terms. 
The whole training process does not rely on any 3D supervision and is fully self-supervised. 



\subsection{3DMM Regression Module}
We employ the 3DMM model including identity and expression parameters:
\begin{equation}
  s=\overline{s}+B_{id}x_{id}+B_{exp}x_{exp},
\label{eq:3dmm}
\end{equation}
where $\overline{s}$ is the vector format of the mean 3D face model, $B_{id}$ and $B_{exp}$ are the identity bases and the expression bases from  \cite{blanz1999morphable} and  \cite{cao2014facewarehouse}, respectively. The 3D reconstruction is formulated as regressing 3DMM parameters $x_{id}$ and $x_{exp}$ in Eq. \eqref{eq:3dmm}.

Our 3DMM regression module employs a similar framework with existing methods \cite{tewari2017mofa, genova2018unsupervised}. It takes a color face image as input and transforms it progressively to the latent code vector using multiple convolutional layers and nonlinear activations. Specifically, we adopt the VGG-Face \cite{Parkhi15vggface} structure in the 3DMM encoder. As we notice the feature representation discrepancies between 2D face images and 3D face models, we randomly initialize the network parameters and train them from scratch. During the training process, we project the output 3D face model into a 2D face image. The loss functions are mainly designed to measure the difference between the projected face and the input face. The total loss function for training the 3DMM-based coarse model is denoted as:
\begin{equation}
    \begin{aligned}
        \mathcal{L}_{\textrm{coarse}} =
        & w_1\cdot\mathcal{L}_{\textrm{pixel}}+w_2\cdot\mathcal{L}_{\textrm{lm}} 
        & +w_3\cdot\mathcal{L}_{\textrm{id}}+w_4\cdot\mathcal{R}_{\textrm{param}},
        \label{eq:loss}
    \end{aligned}
\end{equation}
where $\mathcal{L}_{\textrm{pixel}}$ is the photometric loss, $\mathcal{L}_{\textrm{lm}}$ is the landmark consistency loss, $\mathcal{L}_{\textrm{id}}$ is the perceptual identity loss, and $\mathcal{R}_{\textrm{param}}$ is the 3DMM parameter regularization term. The weights \{$w_1$, $w_2$, $w_3$, $w_4$\} control the influence of each term and are set as constant values. The details are as follows.

\subsubsection{Photometric loss}
The photometric loss is set to measure the pixel-wise difference between the input face image and the rendered face image. We denote the input face image as $I$ and the rendered face image as $I^R$. The loss function is defined as:
\begin{equation}
    \mathcal{L}_{pixel} = \dfrac{1}{|\mathcal{M}|}\sum_{(i,j) \in \mathcal{M}}|| I_{i,j} - I^R_{i,j} ||_{2},
\end{equation}
where $\mathcal{M}$ deontes the visible pixels on the $I^R$ and $(i,j)$ is the location of each visible pixel. We compute the photometric loss by averaging the $L_{2,1}$-distances for all visible pixels.

\subsubsection{Landmark Consistency Loss}
The landmark consistency loss measures the $L2$-distance between the 68 detected landmarks in the input face image and the rendered locations of the 68 key points in the 3D mesh. The detection of 2D landmarks is explained in  \cite{bulat2017far}.
The loss function is defined as:
\begin{equation}
    \mathcal{L}_{lm} = \dfrac{1}{N}\sum_{i = 1}^{N}|| p_{i} - p^R_i ||_{2}^{2},
\end{equation}
where $p_{i}$ is the $i$-th landmark position in the input face image, $p^R_i$ is the corresponding $i$-th landmark position in the rendered face image, and $N=68$ is the number of landmarks. The landmark consistency loss effectively controls the pose and expression of the 3D face model.

\subsubsection{Perceptual Identity Loss}
The perceptual identity loss reflects the perception similarity between two images. We send both the input face image and the rendered face image into the VGG-face recognition network  \cite{Parkhi15vggface} for feature extraction. We denote the extracted CNN features of the input face image as $\phi(I)$, the features of the rendered face image as $\phi(I^R)$. The perceptual consistency loss is defined as:
\begin{equation}
    \mathcal{L}_{id} = || \phi(I) - \phi(I^R) ||_{2}^{2},
\end{equation}
where $\phi$ is the parameters of the VGG-face network  \cite{Parkhi15vggface} and is kept fixed during the training process.

\subsubsection{Regularization Term}
We propose a regularization term for the 3DMM parameters.
Since the values of the 3DMM parameters are subject to normal distribution, we have to prevent their values from deviating from zeros too much. Otherwise, the 3D faces reconstructed from the parameters are distorted. The regularization term is:

\begin{equation}
        \mathcal{R}_{param} =  \omega_{s} ||x_{id}||^2 + \omega_{e} ||x_{exp}||^2,
     \label{eq:reg}
\end{equation}
where $\omega_{s}$ and $\omega_{e}$ are weighting parameters.

\subsection{Detail Modeling Module}

The detail modeling module is an image-to-image translation network in UV space. It consists of an encoder-decoder network with skip connections.
The input image and the reconstructed coarse 3D face model are unwrapped into two UV texture maps of the same resolution.
These two UV maps are concatenated and the invisible regions are masked out based on the estimated 3D poses.
Then, the concatenated UV maps are fed into the encoder-decoder network.
The network produces a displacement depth map.
This map is added to the UV position map of the 3DMM-based coarse model to generate a refined UV map, which is wrapped back to a 2D face image by the UV render layer.
We compare the output 2D image of the detail modeling network and the input image with a pixel-level photometric loss.
%
%
During training, we use smoothness loss and regularization terms together with the photometric loss.
The smoothness loss and regularization terms are set on the displacement depth maps to reduce both artifacts and distortions in the reconstruction process.

The total loss function of the detail modeling network is:
\begin{equation}
    \begin{aligned}
        \mathcal{L}_{\textrm{fine}} =
        & \omega_{p}\cdot\mathcal{L}_{\textrm{pixel}} +  \omega_{s}\cdot\mathcal{L}_{\textrm{smooth}}  + \omega_{d}\cdot\mathcal{L}_{\textrm{disp}},
        \label{eq:loss-detail}
    \end{aligned}
\end{equation}
where $\mathcal{L}_{\textrm{pixel}}$ is the photometric loss to measure the pixel-wise difference between the input image and rendered 2D face image from UV renderer. The $\mathcal{L}_{\textrm{smooth}}$ term is the smoothness loss, and $\mathcal{L}_{\textrm{disp}}$ is the regularization term on displacement map. The weights \{$w_{p}$, $w_{s}$, $w_{d}$ \} are constant values to balance the influence of each loss term. We will introduce the smoothness loss and the regularization terms below:

\subsubsection{Smoothness Loss}
We propose the smoothness loss on both the UV displacement normal map and the displacement depth map to ensure similar representation of the neighboring pixels on these maps. Another advantage of the smoothness loss is that it ensures the robustness to mild occlusions. The smoothness loss can be written as:
\begin{multline}
    \mathcal{L}_{\textrm{smooth}} = \sum_{i \in \mathcal{V}_{UV}} \sum_{j \in \mathcal{N}(i)}w_{sn} || \Delta n(i) - \Delta n(j) || ^2 \\
     + w_{sz} || \Delta z(i) - \Delta z(j) ||^2,
\end{multline}
where $\Delta n(i)$ is the difference measurement on pixel $i$ in the UV map. It computes the pixel distance between the original UV normal map (i.e., the vertex normal computed from coarse 3D model) and the UV normal map integrated with the displacement depth map. Similarly, $\Delta z(i)$ computes the pixel distance between the original displacement depth map and the updated displacement depth map. The $\mathcal{V}_{UV}$ are vertices in the UV space and $\mathcal{N}(i)$ is the neighborhood of vertex $i$ with a radius of 1. The $\Delta n(i)$ measures the difference between the UV normal map before and after adding displacement map. The weights $w_{sn}$ and $w_{sz}$ are used to combine these two smoothing losses and they are set as 20 and 10.

\subsubsection{Regularization Term}
We propose the regularization terms on both the displacement depth map and the displacement normal map to reduce severe depth changes, which may introduce distortion in face on the 3D mesh. The regularization term can be written as:
\begin{equation}
    \mathcal{L}_{\textrm{disp}} = \sum_i  w_{dn}  || \Delta n(i) || ^2 + w_{dz} || \Delta z(i) ||^2,
\end{equation}
where $w_{dn}$ and $w_{dz}$ are set to $0.5$ and $0.01$, respectively.

\subsection{Camera View}
The pose parameter in the proposed model is $7$D, including scale $f$, rotation angles(in rads) $r_x, r_y, r_z$, and translation $t_x, t_y, t_z$. We apply orthogonal projection to project the 3D vertices into 2D. We denote the vertex in 3D as $\mathbf{v}$, the projection operation as $\Pi$, the projected 2D points as $\mathbf{p}$, respectively. Then we have:

\begin{equation}
	\mathbf{p} = \Pi(f\mathbf{R}\mathbf{v} + \mathbf{t}),
\end{equation}
where $\mathbf{R}$ is the rotation matrix computed by $r_x, r_y, r_z$, and $\mathbf{t} = (t_x, t_y,t_z)^T$ is the translation vector.

\subsection{Rendering Layer}

The rendering layer is a modification to  \cite{genova2018unsupervised}. We use spherical harmonics as our lighting model instead of the Phong reflection model. And orthogonal projection is applied here.

\subsection{UV Render Layer}

The UV render layer takes two inputs. One is a coarse UV position map that is built by unwrapping the coarse face mesh. The other is a predicted displacement depth map in UV space from the detailed modeling module. 
Using these two inputs, the UV render layer first computes the detail UV position map by adding the displacement to the coarse UV position map. 
Then a final output triangle mesh can be generated by connecting neighboring pixels in the detail UV position map. 
With the pose, lighting and texture parameters estimated in 3DMM regression module, a final image can be rendered. 
The whole rendering process is differentiable. 
For implementation details, please refer to our code at: \url{https://github.com/cyj907/unsupervised-detail-layer}. 


\section{Experiments}
In this section, we provide more implementation details and conduct extensive experiments to demonstrate the effectiveness of our method. 

\begin{figure*}[htbp]
    \centering
    \setlength{\tabcolsep}{0.01em} 
\clearpage
\onecolumn
\footnotesize
  \begin{longtable}[t]{ccccccccc}
  	\multirow{2}{*}{\textbf{Input}} & \multirow{2}{*}{\textbf{MoFA \cite{tewari2017mofa}}} & \multirow{2}{*}{\textbf{Ours-3DMM}} & \multicolumn{2}{c}{\textbf{MoFA \cite{tewari2017mofa}}} & \multicolumn{2}{c}{\textbf{Genova18 \cite{genova2018unsupervised}}} & \multicolumn{2}{c}{\textbf{Ours-3DMM}}\\
	& & & \textbf{Shape} & \textbf{Texture} & \textbf{Shape} & \textbf{Texture} & \textbf{Shape} & \textbf{Texture}\\
    \endfirsthead
    \endhead
    \endfoot
    \endlastfoot
    \includegraphics[width=0.1\linewidth]{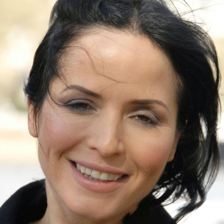} &
	\includegraphics[width=0.1\linewidth]{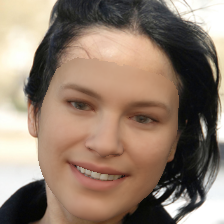} &
	\includegraphics[width=0.1\linewidth]{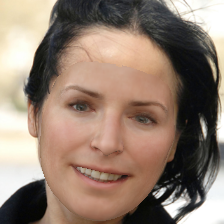} &
	\includegraphics[width=0.1\linewidth]{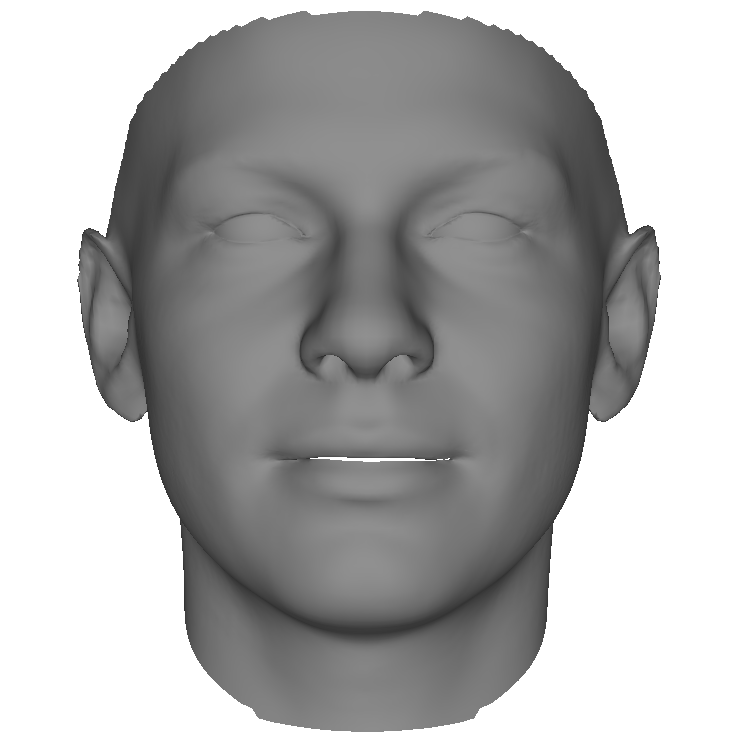} &
	\includegraphics[width=0.1\linewidth]{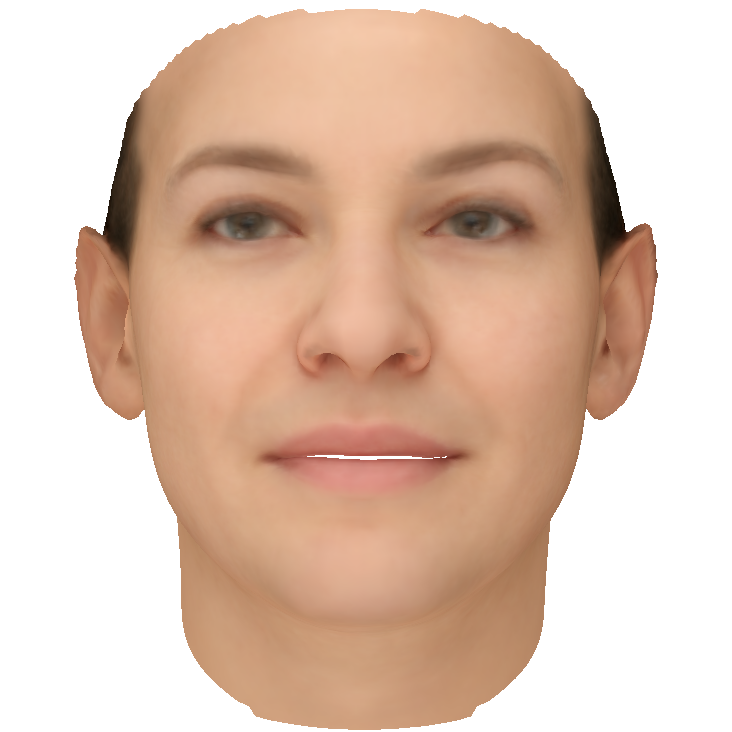} &
	\includegraphics[width=0.1\linewidth]{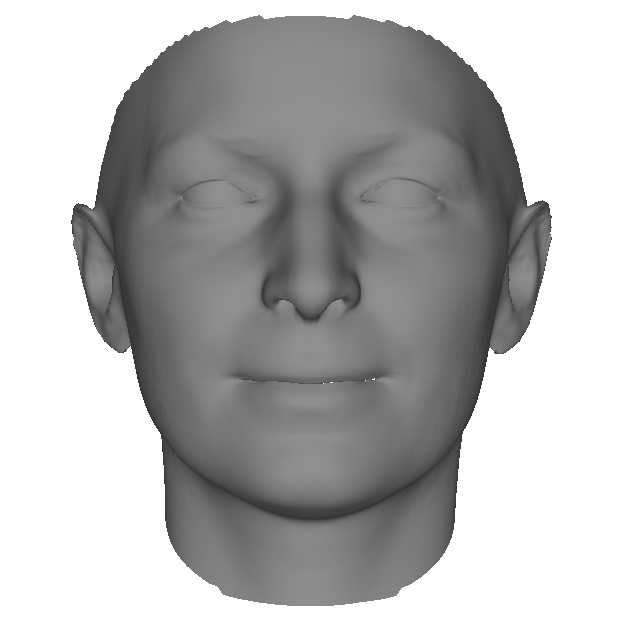} &
	\includegraphics[width=0.1\linewidth]{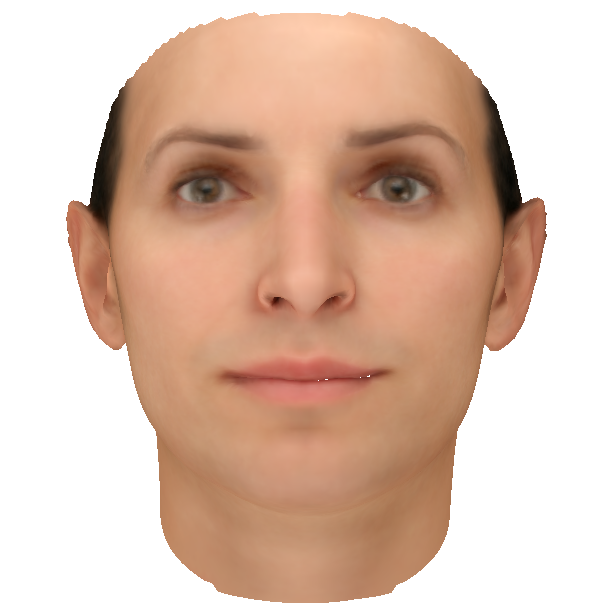} &
	\includegraphics[width=0.1\linewidth]{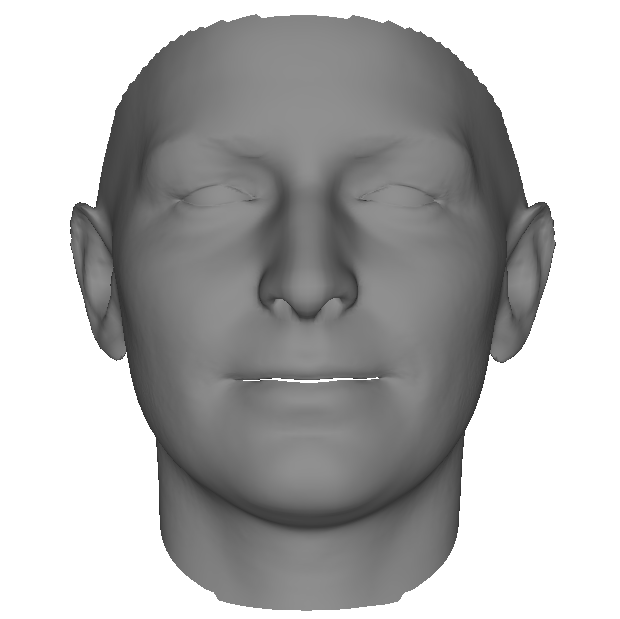} &
	\includegraphics[width=0.1\linewidth]{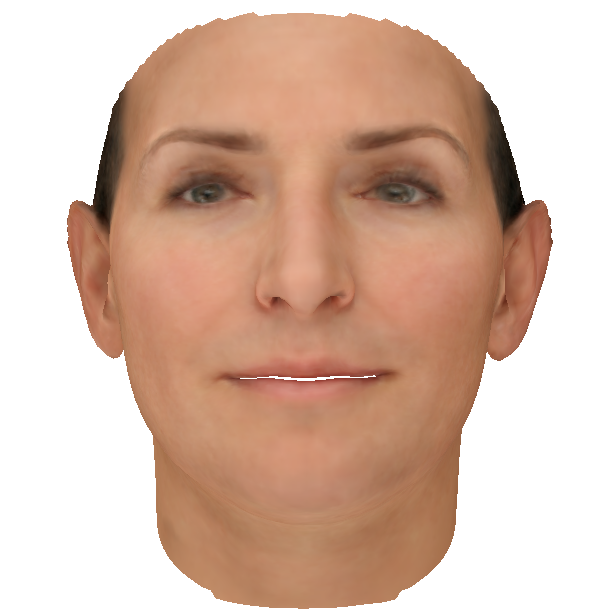}
	\\
	
	\includegraphics[width=0.1\linewidth]{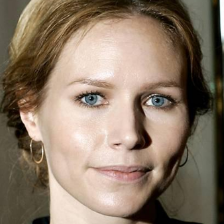} &
	\includegraphics[width=0.1\linewidth]{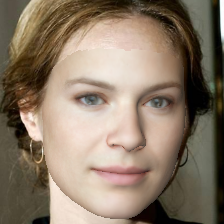} &
	\includegraphics[width=0.1\linewidth]{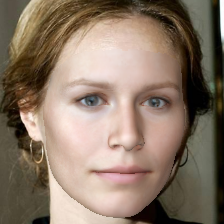} &
	\includegraphics[width=0.1\linewidth]{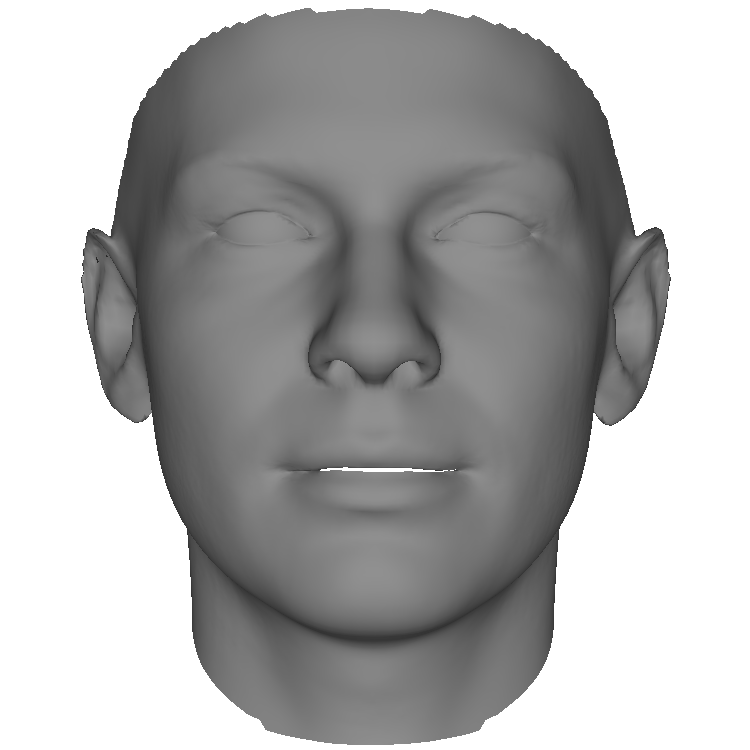} &
	\includegraphics[width=0.1\linewidth]{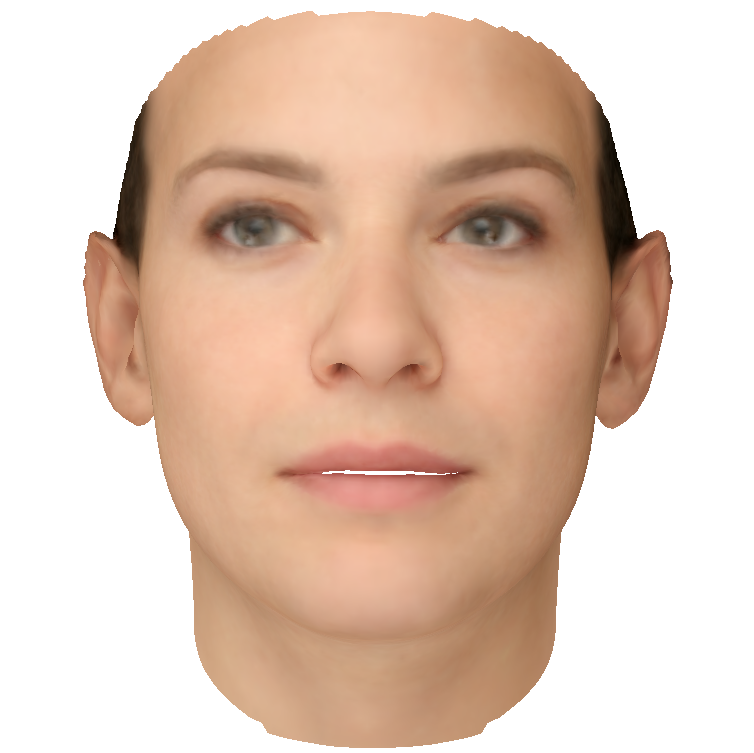} &
	\includegraphics[width=0.1\linewidth]{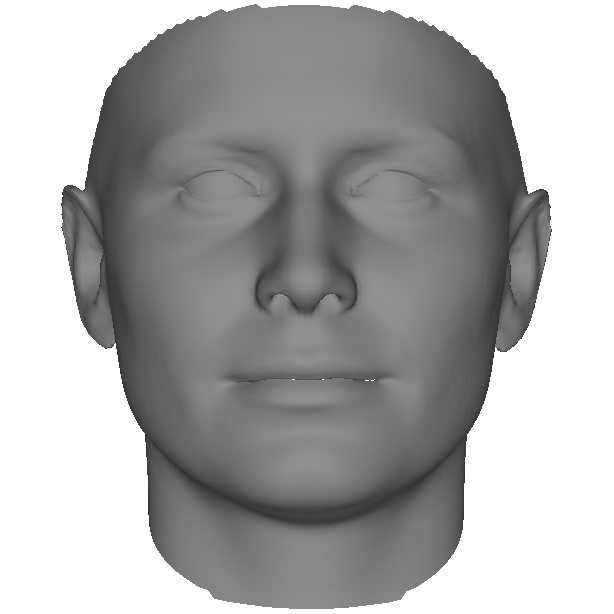} &
	\includegraphics[width=0.1\linewidth]{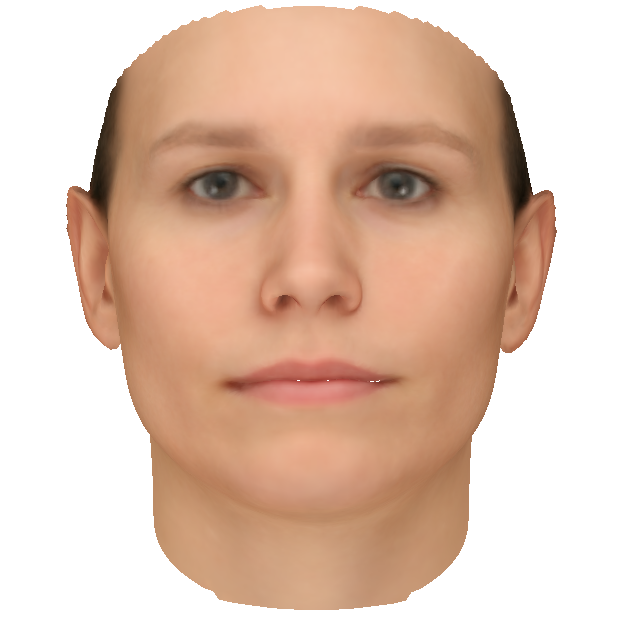} &
	\includegraphics[width=0.1\linewidth]{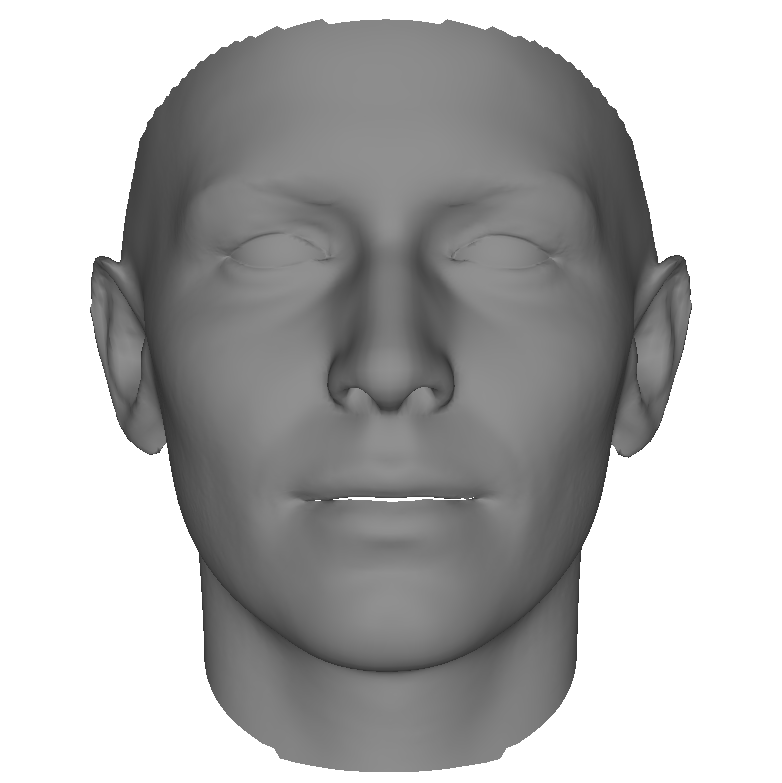} &
	\includegraphics[width=0.1\linewidth]{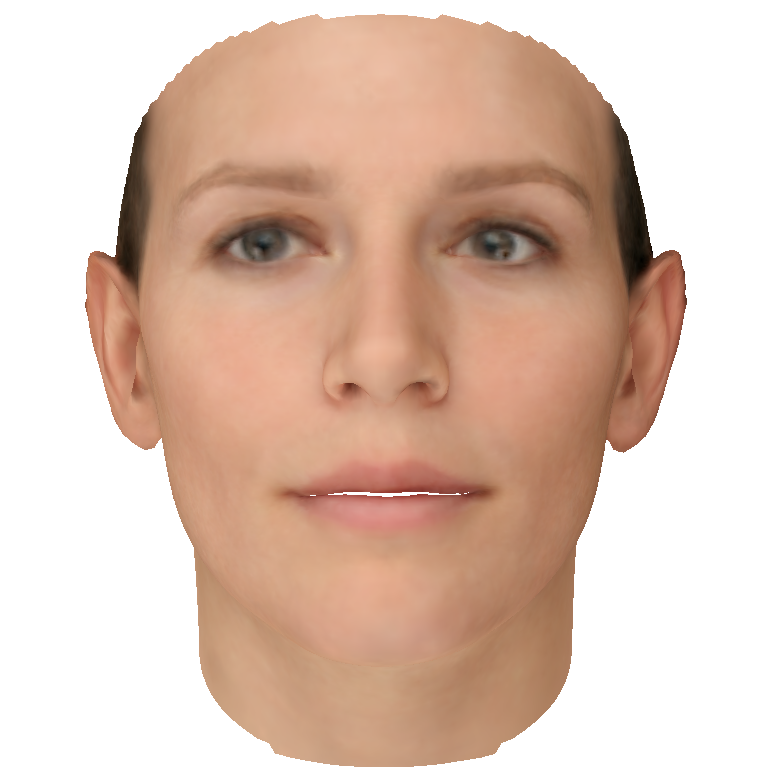}
	\\
	
	
	\includegraphics[width=0.1\linewidth]{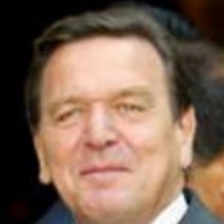} &
	\includegraphics[width=0.1\linewidth]{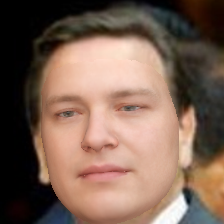} &
	\includegraphics[width=0.1\linewidth]{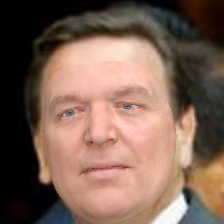} &
	\includegraphics[width=0.1\linewidth]{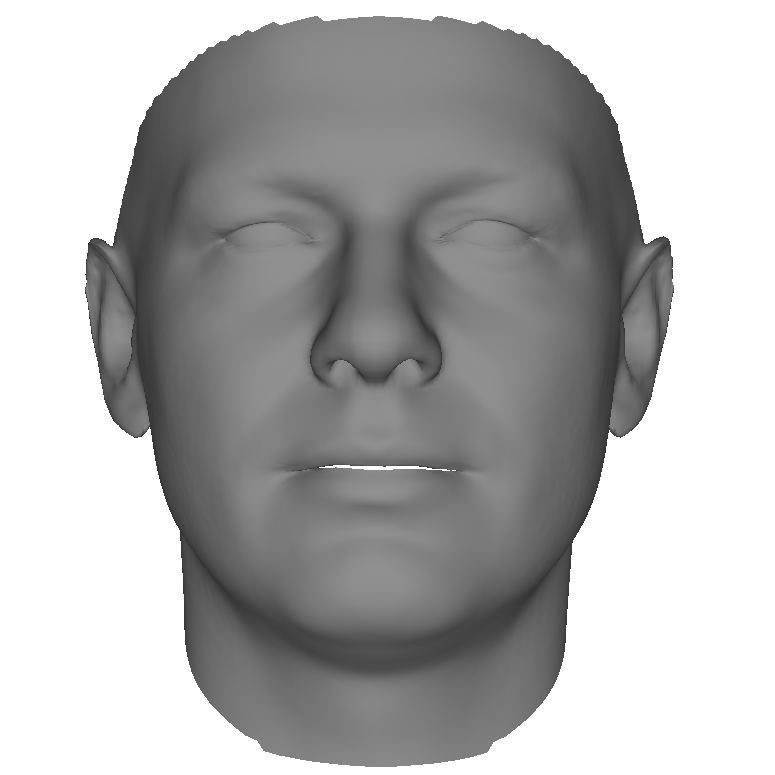} &
	\includegraphics[width=0.1\linewidth]{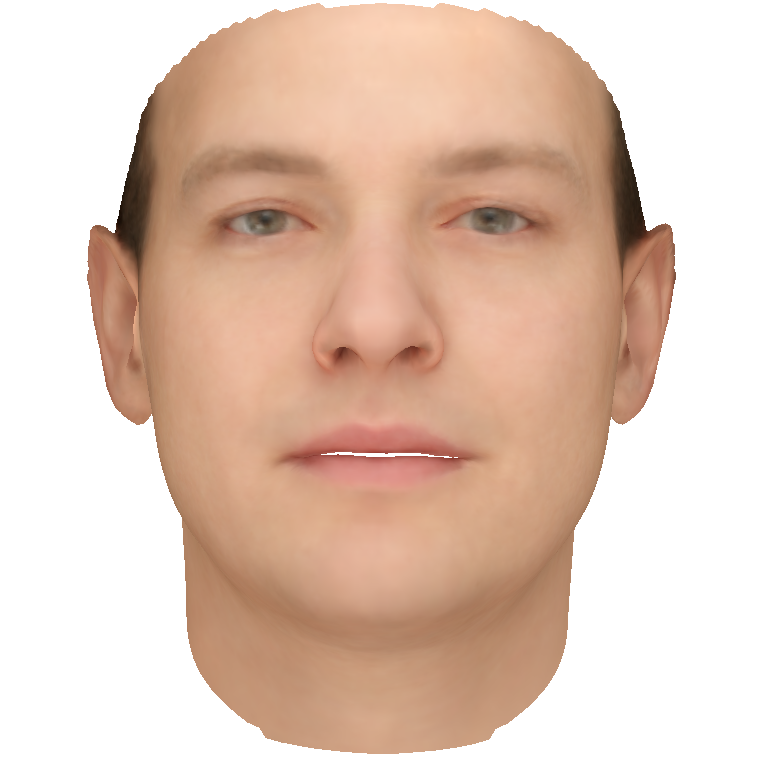} &
	\includegraphics[width=0.1\linewidth]{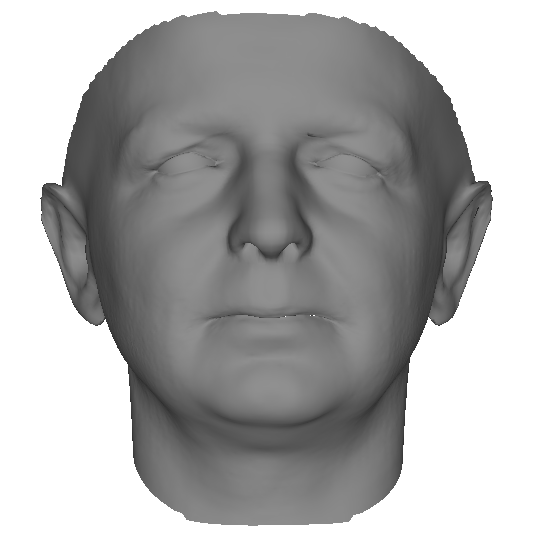} &
	\includegraphics[width=0.1\linewidth]{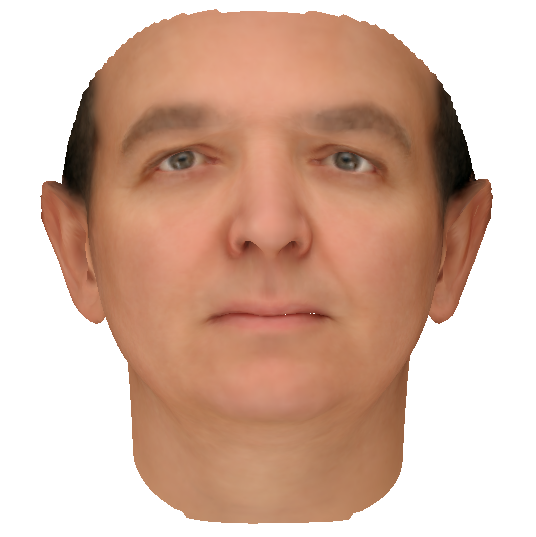} &
	\includegraphics[width=0.1\linewidth]{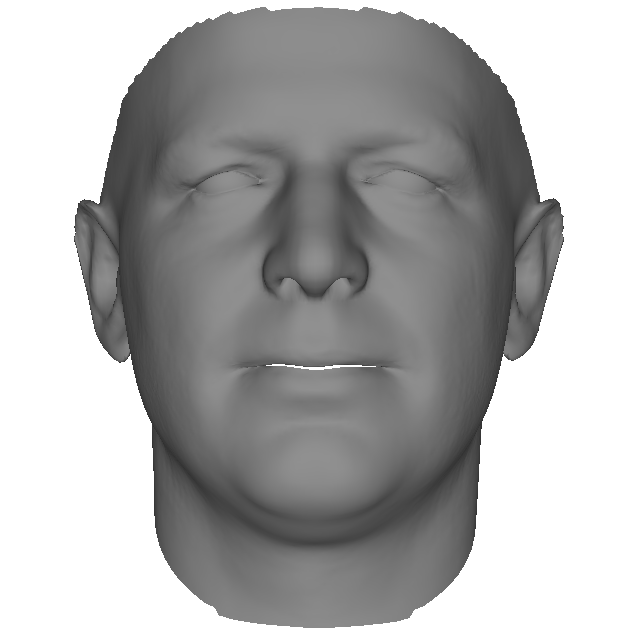} &
	\includegraphics[width=0.1\linewidth]{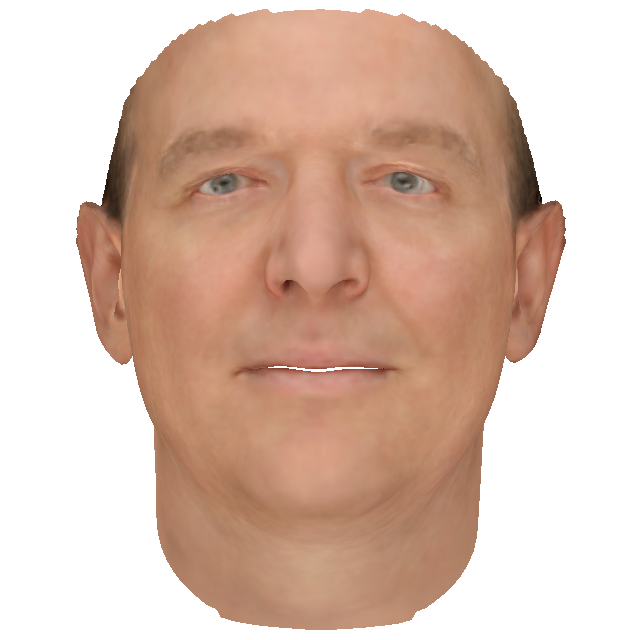}
	\\
	
	\includegraphics[width=0.1\linewidth]{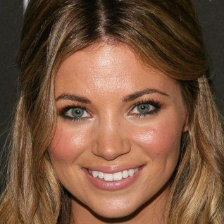} &
	\includegraphics[width=0.1\linewidth]{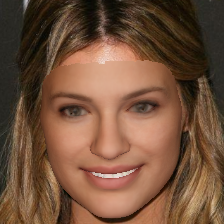} &
	\includegraphics[width=0.1\linewidth]{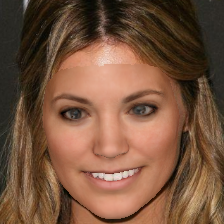} &
	\includegraphics[width=0.1\linewidth]{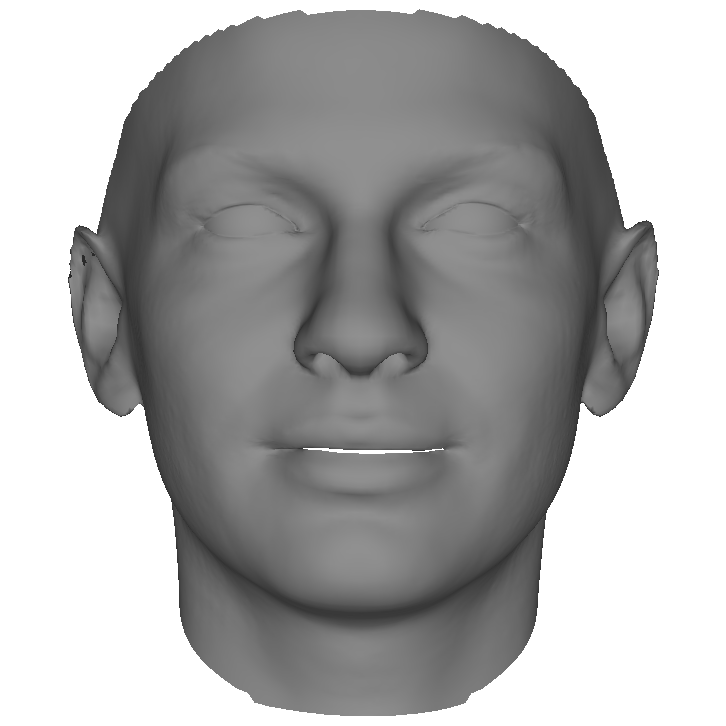} &
	\includegraphics[width=0.1\linewidth]{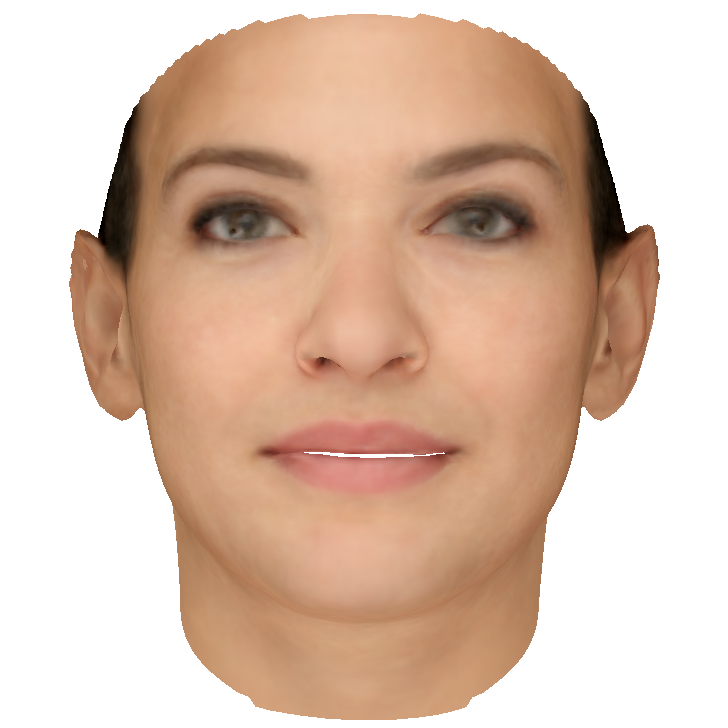} &
	\includegraphics[width=0.1\linewidth]{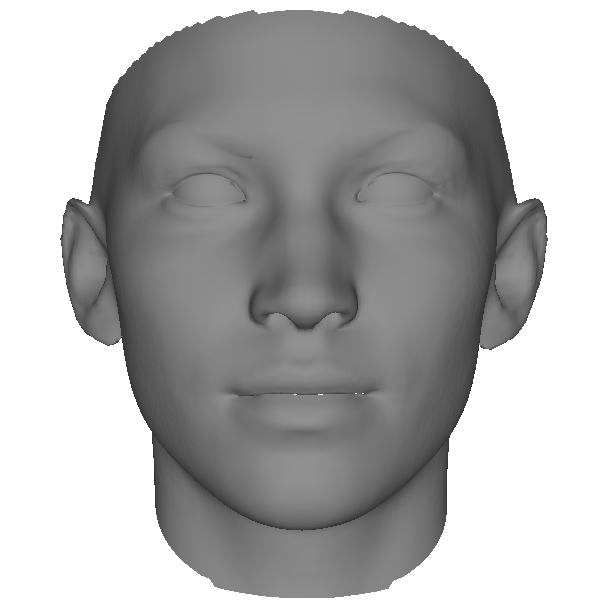} &
	\includegraphics[width=0.1\linewidth]{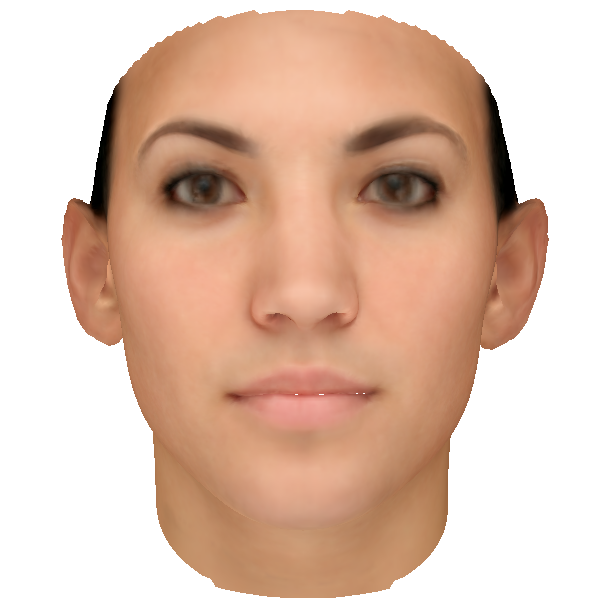} &
	\includegraphics[width=0.1\linewidth]{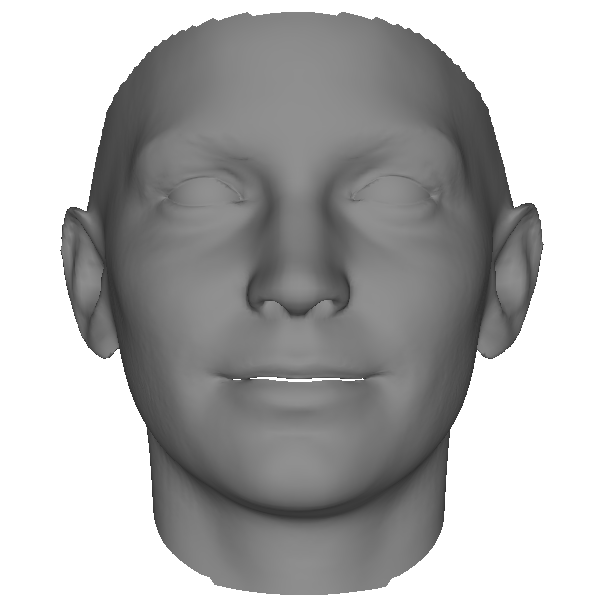} &
	\includegraphics[width=0.1\linewidth]{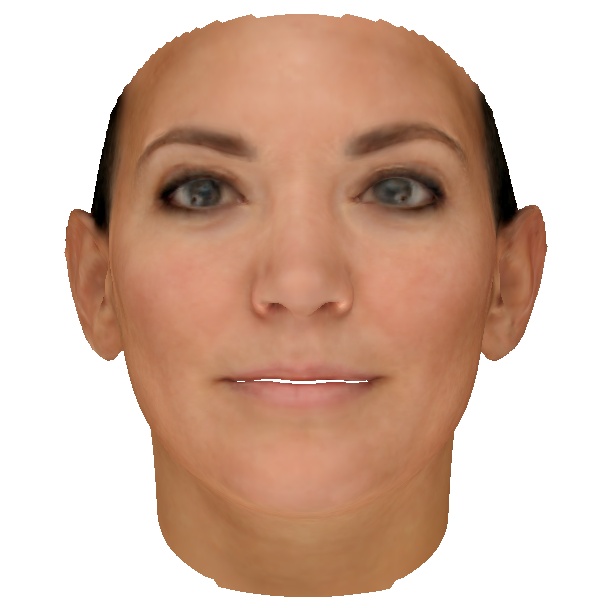}
	\\

	\includegraphics[width=0.1\linewidth]{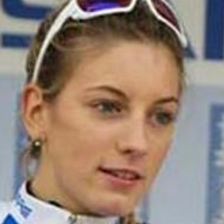} &
	\includegraphics[width=0.1\linewidth]{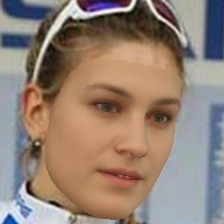} &
	\includegraphics[width=0.1\linewidth]{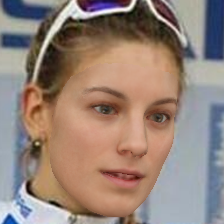} &
	\includegraphics[width=0.1\linewidth]{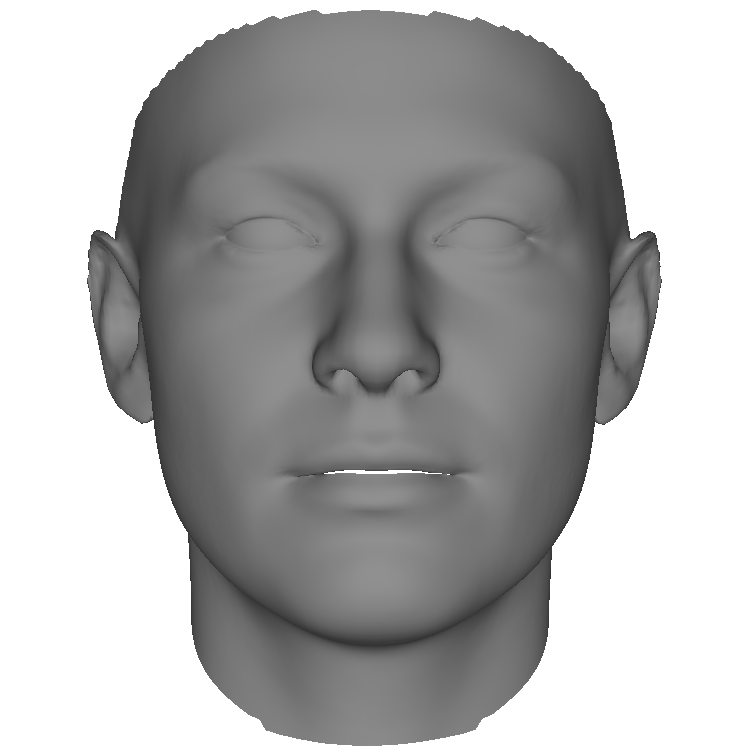} &
	\includegraphics[width=0.1\linewidth]{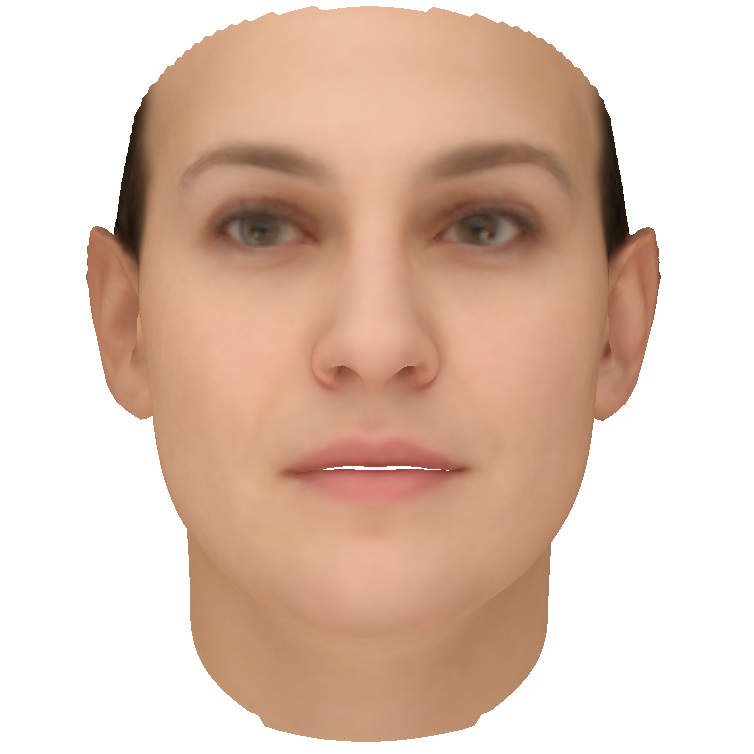} &
	\includegraphics[width=0.1\linewidth]{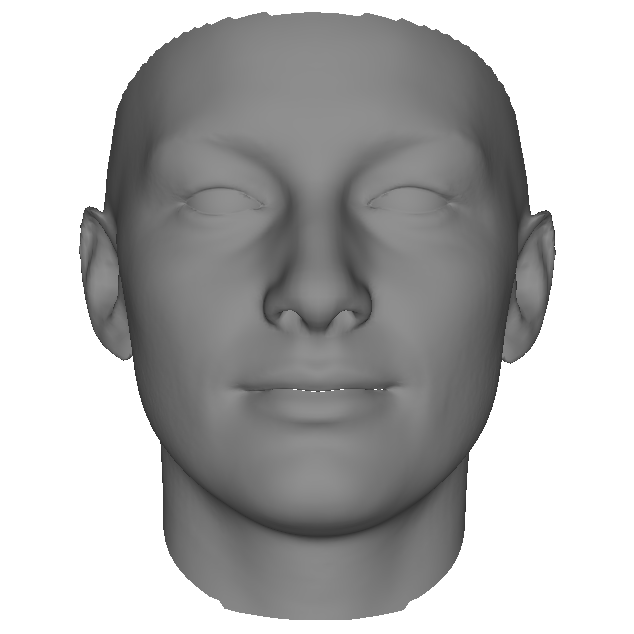} &
	\includegraphics[width=0.1\linewidth]{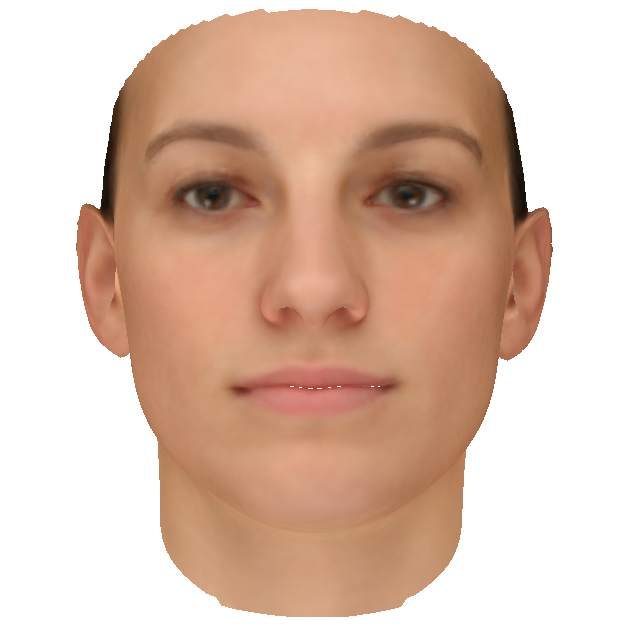} &
	\includegraphics[width=0.1\linewidth]{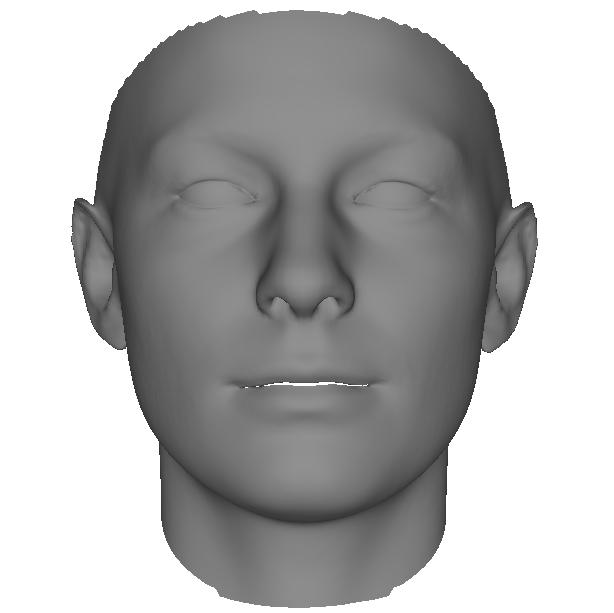} &
	\includegraphics[width=0.1\linewidth]{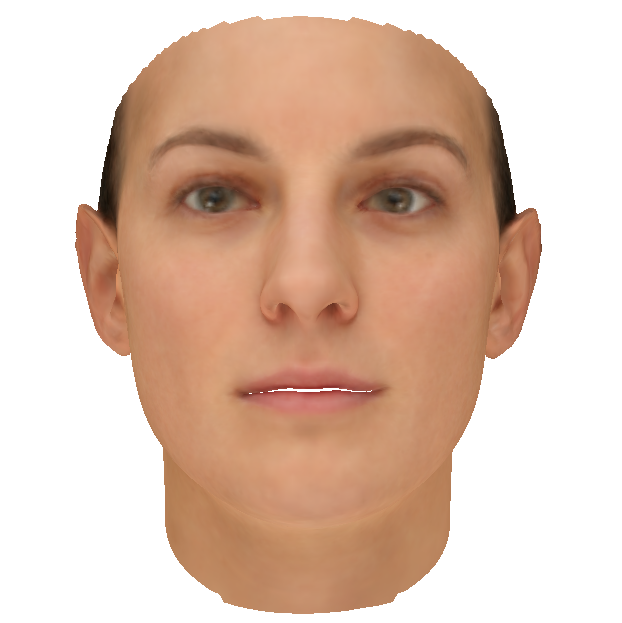}
	\\
	
	
	\includegraphics[width=0.1\linewidth]{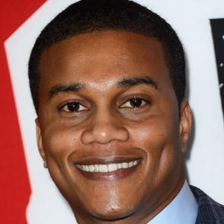} &
	\includegraphics[width=0.1\linewidth]{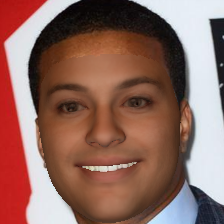} &
	\includegraphics[width=0.1\linewidth]{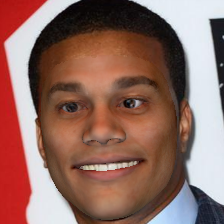} &
	\includegraphics[width=0.1\linewidth]{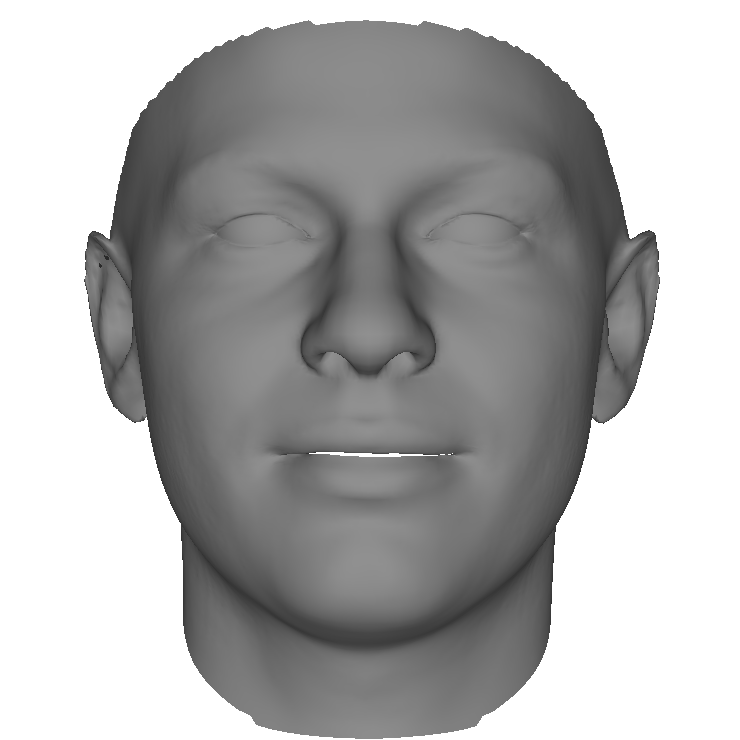} &
	\includegraphics[width=0.1\linewidth]{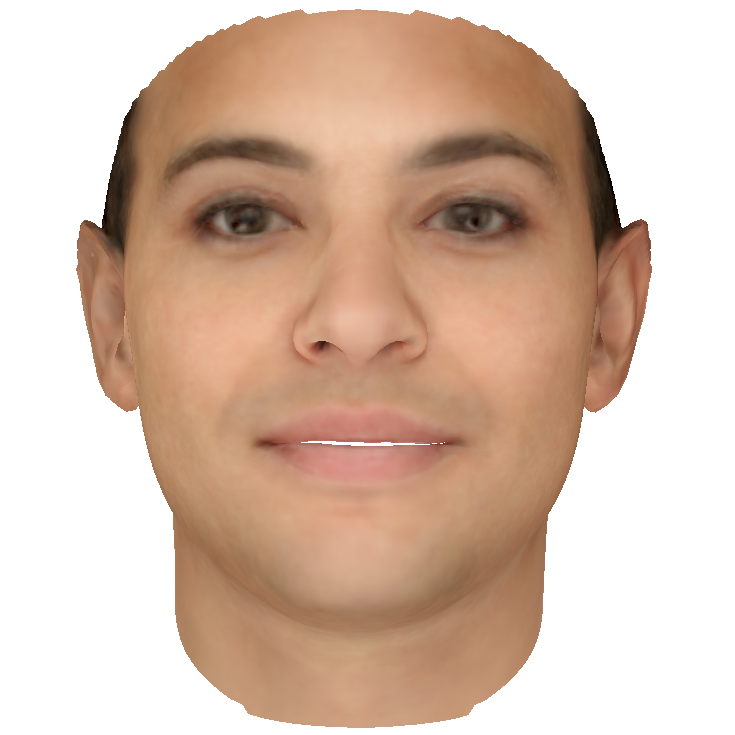} &
	\includegraphics[width=0.1\linewidth]{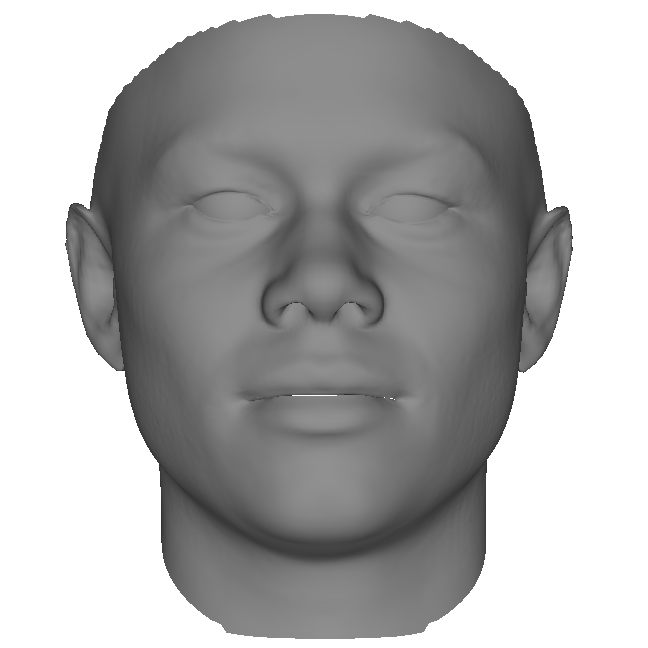} &
	\includegraphics[width=0.1\linewidth]{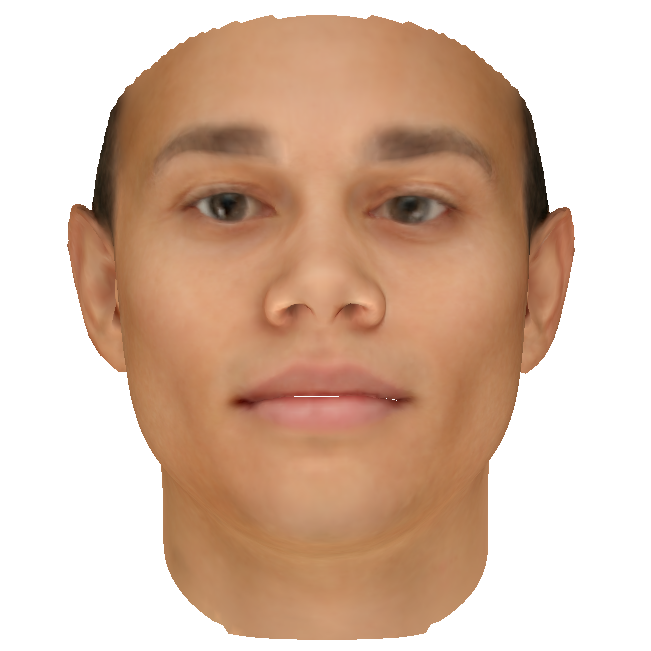}  &
	\includegraphics[width=0.1\linewidth]{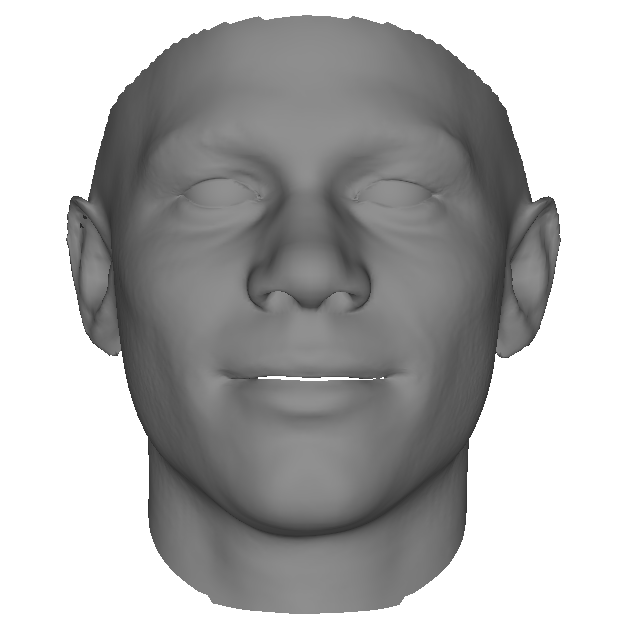} &
	\includegraphics[width=0.1\linewidth]{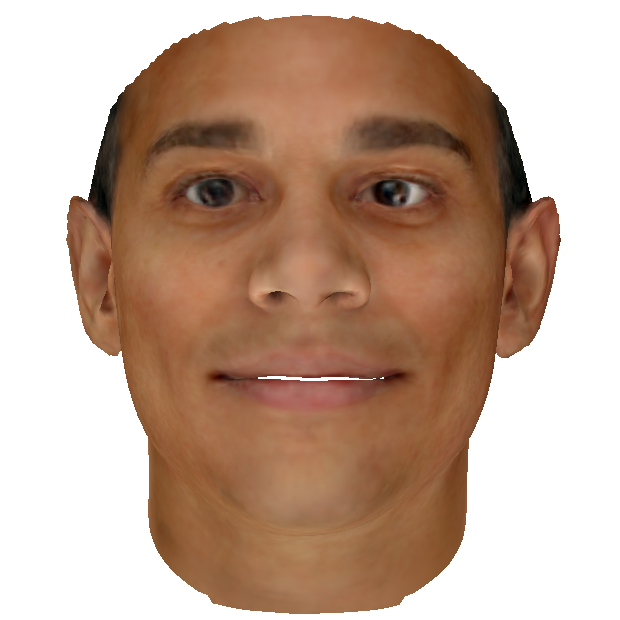}
	\\

    \end{longtable}
    \vspace{-1mm}
    \caption{Results of 3DMM regression on the CelebA \cite{liu2015celeba} and LFW \cite{LFWTech} datasets. For a fair comparison to Genova18 \cite{genova2018unsupervised}, The facial  expressions obtained by MoFA \cite{tewari2017mofa} and our method are set to neutral.}
    \label{fig:eval}
    \clearpage
    \twocolumn
\end{figure*}

\subsection{Implementation Details}


We train our model on the CelebA dataset  \cite{liu2015celeba}.
Before training, we use a landmark detector  \cite{bulat2017far} to exclude failure samples that are not faces.
Then, we separate the remaining images into two parts.
The first part is the training dataset which contains 162,129 images, out of which we keep 1,000 images for validation.
The second part is the testing set which contains 19,899 images.
The architecture of the 3DMM regression module is VGG-Face  \cite{Parkhi15vggface}.
For the detail modeling module, we employ an image-to-image translation architecture similar to pix2pix \cite{isola2017image}.
We randomly initialize all the weights in our model and train them from scratch.

%
The weighting parameters \{$\omega_{1}, \omega_{2}, \omega_{3}, \omega_{4}$\} controlling the total loss for the coarse model in Eq. \eqref{eq:loss} are set as \{$1.3$, $1.0$, $1.5$, $20.0$\}.
We set the weights \{$\omega_{s}$, $\omega_{e}$, $\omega_{t}$\} in the regularization terms in Eq. \eqref{eq:reg} as \{$1.3$, $1.0$, $1.3$\}.
The weights \{$\omega_{p}, \omega_{s}, \omega_{d}$\} of the total loss for the detail modeling module in Eq. \eqref{eq:loss-detail} are set as \{$1.0$, $10.0$, $10.0$\}.
The training in the first stage for the coarse model uses an initial learning rate as $0.0001$, decaying every $5000$ steps at rate $0.9$.
The learning rate for training the detail modeling module in the second stage is set to $0.002$, decaying every $5000$ steps at rate $0.98$.
The batch size is set to be $10$.
We adopt Adam optimizer to train the network on NVIDIA Tesla M40 for over $200,000$ steps for the coarse model and $20,000$ steps for the detail model.

\subsection{Evaluation on 3DMM Regression}

\subsubsection{Shape Analysis}

\begin{table}[htbp]
\begin{center}
\begin{tabular}{l|ccc}
\hline
\multirow{2}{*}{\textbf{Method}} &  \multicolumn{3}{c}{\textbf{Condition}} \\
& Indoor Cooperative & PTZ Indoor & PTZ Outdoor \\
\hline\hline
MoFA  \cite{tewari2017mofa} &  $1.38\pm0.35$ & $1.27\pm0.29$ & $1.28\pm0.27$ \\
Genova18  \cite{genova2018unsupervised} &  $1.41\pm0.37$ & $1.34\pm0.37$ & $1.26\pm0.31$\\
Ours-3DMM &  $\mathbf{1.35\pm0.31}$ &  $\mathbf{1.27\pm0.24}$  &  $\mathbf{1.25\pm0.21}$\\
\hline
\end{tabular}
\end{center}
\caption{Point-to-plane error on the MICC Florence dataset \cite{Bagdanov2011MICC} for 3DMM-based self-supervised learning approaches.}
\label{tab:micc}
\end{table}

We evaluate the accuracy of the 3DMM regression of the shap on the MICC Florence dataset \cite{Bagdanov2011MICC}.
In this dataset, videos are taken on 53 subjects under three different conditions.
These three conditions are defined as Indoor Cooperative, PTZ Indoor and PTZ Outdoor.
The ground truth 3D scans are provided for 52 out of the 53 people.
We used each video frame as the network input.
Before evaluation, we remove the frames where the faces are not detected by the landmark detector  \cite{bulat2017far}.
The 3D shape model for each video sequence is obtained using the average of the 3DMM shape parameters in the remaining video frames.
We follow the procedures mentioned in  \cite{genova2018unsupervised} to compute the point-to-plane error between the predicted 3D face models and ground truth scans.
Table \ref{tab:micc} lists the comparison of our results to two state-of-the-art 3DMM-based self-supervised learning approaches \cite{tewari2017mofa,genova2018unsupervised}.
Our method outperforms existing methods by combining low-level photometric loss and high-level perceptual identity loss. 
%

We further show some visual comparisons of the results in Fig. \ref{fig:eval} on two other datasets, the CelebA \cite{liu2015celeba} and the LFW \cite{LFWTech} datasets. 
Compared with MoFA \cite{tewari2017mofa} and Genova18  \cite{genova2018unsupervised}, our method is able to generate more faithful results. 
Note that our shape results captures more personalized facial characteristics compared to MoFA, while are more faithful than Genova18. 
For example, the generated face by Genova18 \cite{genova2018unsupervised} in row 3 in Fig. \ref{fig:eval} is too short along the vertical axis compared with the corresponding input image, while our result is more faithful.
When focusing on the texture, we notice that MoFA \cite{tewari2017mofa} tends to generate smooth texture. The results from Genova18 \cite{genova2018unsupervised} are more realistic but does not show sufficient color distinction between people from different races. In contrast, our method shows more color diversity for individuals.


%


\subsubsection{Expression Analysis}

\begin{figure}[t]
    \centering
    \footnotesize
    \setlength{\tabcolsep}{0.1em}
    \begin{tabular}{cccccc}
        Input & MoFA \cite{tewari2017mofa} & Ours-3DMM & Input & MoFA \cite{tewari2017mofa} & Ours-3DMM \\

        \includegraphics[width=0.16\linewidth]{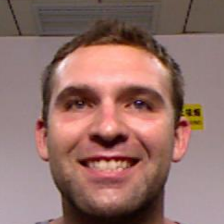} &
        \includegraphics[width=0.16\linewidth]{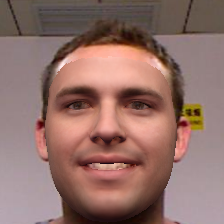} &
        \includegraphics[width=0.16\linewidth]{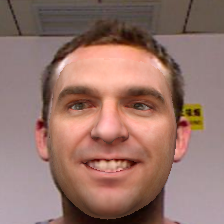} &
        \includegraphics[width=0.16\linewidth]{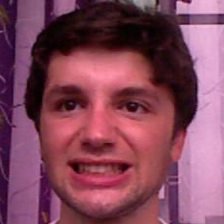} &
        \includegraphics[width=0.16\linewidth]{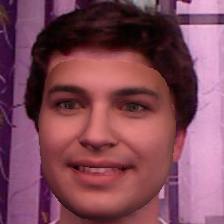} &
        \includegraphics[width=0.16\linewidth]{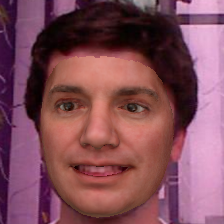} \\

        \includegraphics[width=0.16\linewidth]{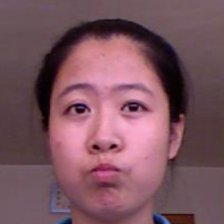} &
        \includegraphics[width=0.16\linewidth]{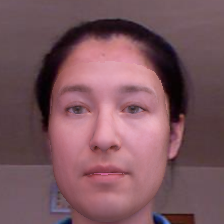} &
        \includegraphics[width=0.16\linewidth]{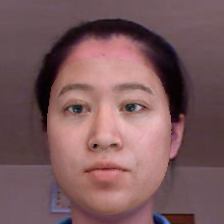} &
        \includegraphics[width=0.16\linewidth]{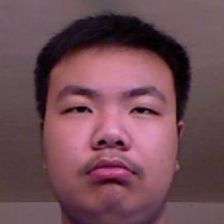} &
        \includegraphics[width=0.16\linewidth]{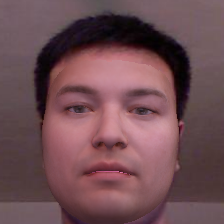} &
        \includegraphics[width=0.16\linewidth]{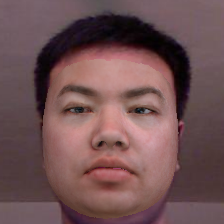}\\

        \includegraphics[width=0.16\linewidth]{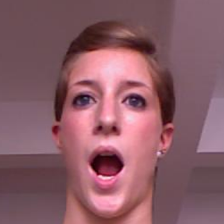} &
        \includegraphics[width=0.16\linewidth]{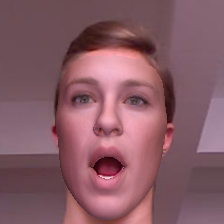} &
        \includegraphics[width=0.16\linewidth]{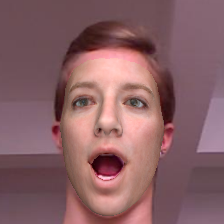} &
        \includegraphics[width=0.16\linewidth]{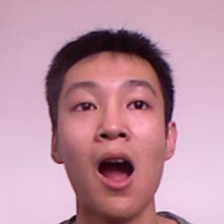} &
        \includegraphics[width=0.16\linewidth]{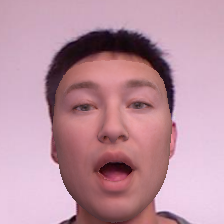} &
        \includegraphics[width=0.16\linewidth]{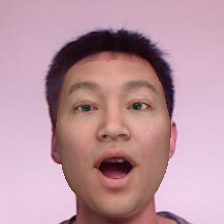} \\

        \includegraphics[width=0.16\linewidth]{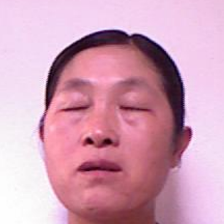} &
        \includegraphics[width=0.16\linewidth]{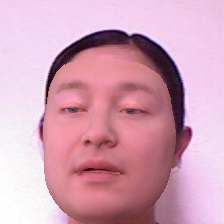} &
        \includegraphics[width=0.16\linewidth]{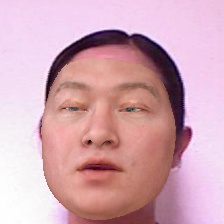} &
        \includegraphics[width=0.16\linewidth]{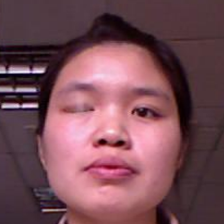} &
        \includegraphics[width=0.16\linewidth]{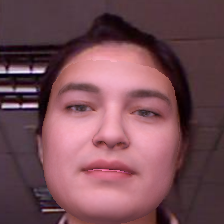} &
        \includegraphics[width=0.16\linewidth]{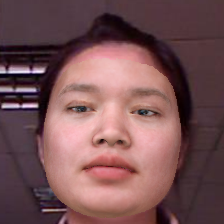} \\

    \end{tabular}
    \caption{3DMM regression results of facial expressions on the FaceWarehouse dataset \cite{cao2014facewarehouse} for MoFA \cite{tewari2017mofa} and our method.}
    \label{fig:exp}
\end{figure}

\begin{table}[t]
\begin{center}
\begin{tabular}{c|c|c}
\hline
\textbf{Method} & \textbf{MoFA \cite{tewari2017mofa}} & \textbf{Ours} \\
\hline\hline
Error & $2.26\pm0.58$ & $\mathbf{1.81\pm0.43}$ \\
\hline
\end{tabular}
\end{center}
\caption{Mean and standard deviation of point-to-point RMSE on FaceWarehouse \cite{cao2014facewarehouse} for 3DMM regression with expressions.}
\label{tab:exp}
\end{table}

We evaluate the expression reconstruction results on the FaceWarehouse dataset \cite{cao2014facewarehouse}.
The dataset contains 150 subjects in 20 different expressions.
We compare our method with MoFA \cite{tewari2017mofa}.
We first use non-rigid registration to compute the vertex correspondence between FaceWarehouse data and 3DMM model.
Then, we apply rigid transforms to align the predicted meshes and the ground truth scans provided in FaceWarehouse \cite{cao2014facewarehouse}.
We compute the point-to-point RMSE errors (root-mean-square-error) for the corresponding vertices of the two meshes.
Table \ref{tab:exp} shows the mean and standard deviation of the point-to-point RMSE error on the FaceWarehouse \cite{cao2014facewarehouse} dataset.
Since Genova18 \cite{genova2018unsupervised} does not estimate expression parameters, we only compare to MoFA \cite{tewari2017mofa}.
Our method produces more accurate results than MoFA.
Fig. \ref{fig:exp} shows some visual results of the two methods.
%
When the commonly-seen expression (e.g., smiling with visible teeth) appears on the subjects of the input image as shown in the first row, both MoFA and our method are effective to reconstruct the 3D model.
Meanwhile, the 3D model generated by our method contains more identity-specific details.
When some uncommon expression appears (e.g., pouted mouths) on the second row, MoFA does not reconstruct the 3D model effectively while our method does.
When the expression is extreme (e.g., largely-open mouths and closed eyes) as shown in the last row, neither of these two methods performs well. However, our method still performs favorably against MoFA.
%


\begin{figure}[t]
    \centering
    \footnotesize
    \setlength{\tabcolsep}{0.01em}
    \begin{tabular}{ccccccc}
        \multirow{2}{*}{\textbf{Input}} & \multicolumn{3}{c}{\textbf{MoFA  \cite{tewari2017mofa}}} & \multicolumn{3}{c}{\textbf{Ours-3DMM}} \\
         & Overlay & Light & Albedo & Overlay & Light & Albedo \\
        \includegraphics[width=0.14\linewidth]{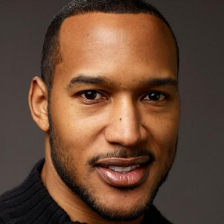} &

        \includegraphics[width=0.14\linewidth]{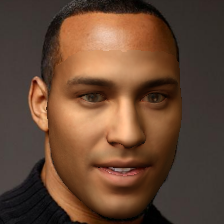} &
        \includegraphics[width=0.14\linewidth]{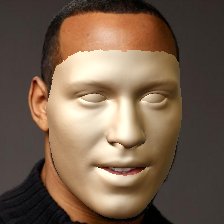} &
        \includegraphics[width=0.14\linewidth]{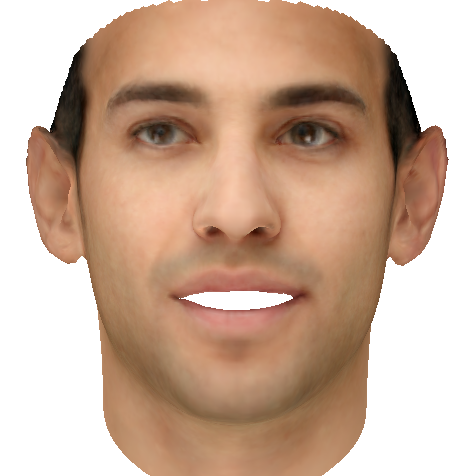} &

        \includegraphics[width=0.14\linewidth]{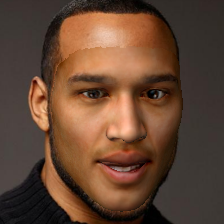} &
        \includegraphics[width=0.14\linewidth]{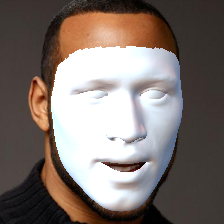} &
        \includegraphics[width=0.14\linewidth]{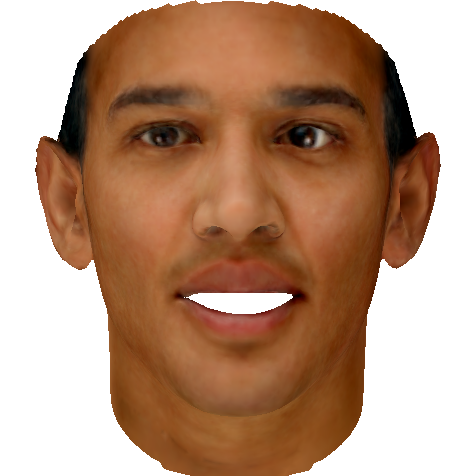} \\

        \includegraphics[width=0.14\linewidth]{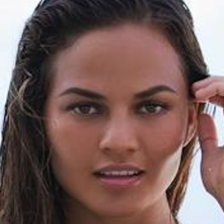} &
        \includegraphics[width=0.14\linewidth]{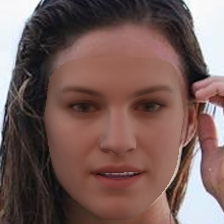} &
        \includegraphics[width=0.14\linewidth]{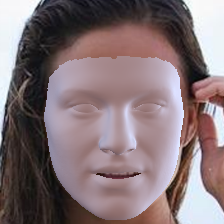} &
        \includegraphics[width=0.14\linewidth]{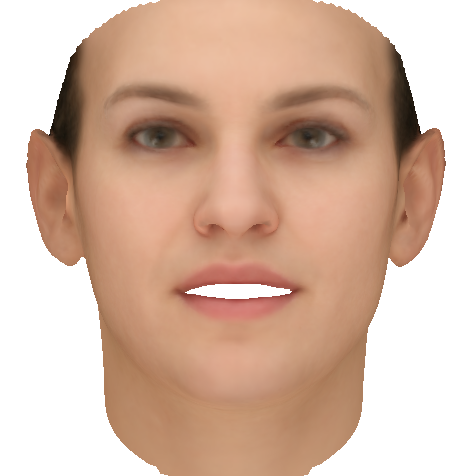} &
        \includegraphics[width=0.14\linewidth]{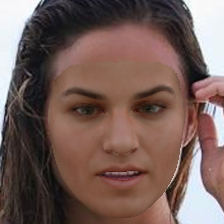} &
        \includegraphics[width=0.14\linewidth]{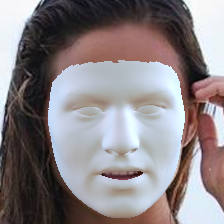} &
        \includegraphics[width=0.14\linewidth]{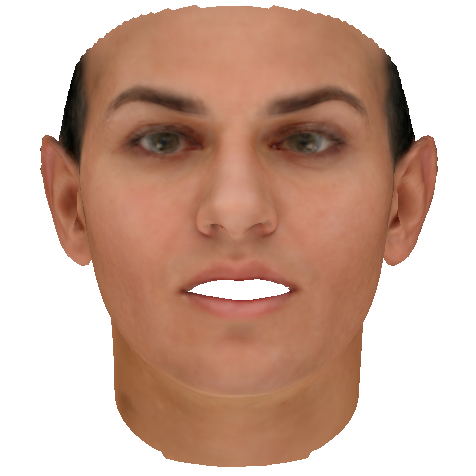} \\

        \includegraphics[width=0.14\linewidth]{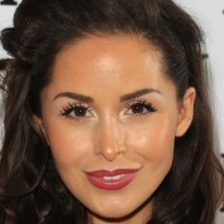} &
        \includegraphics[width=0.14\linewidth]{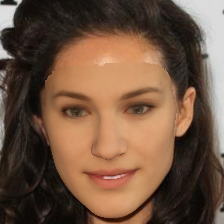} &
        \includegraphics[width=0.14\linewidth]{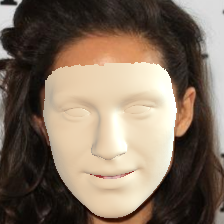} &
        \includegraphics[width=0.14\linewidth]{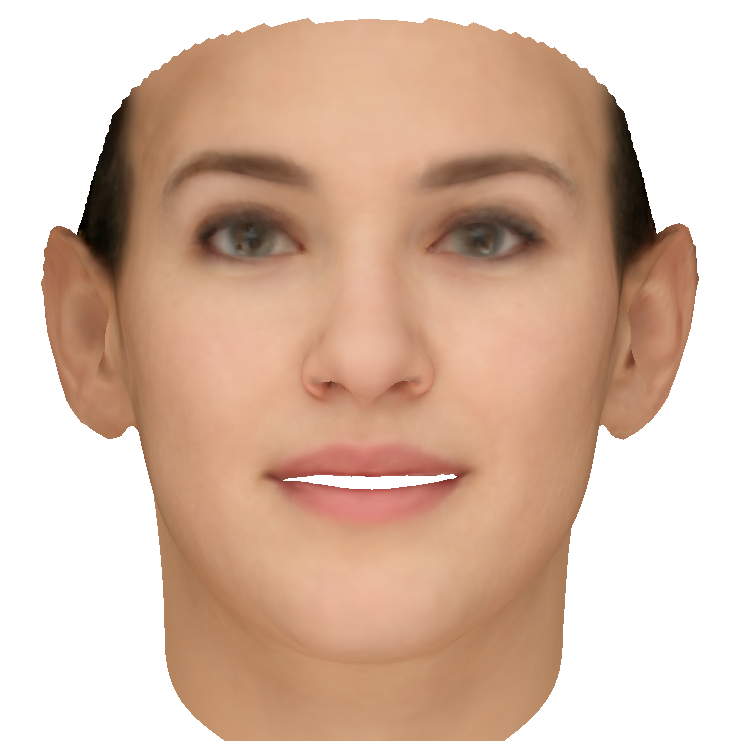} &
        \includegraphics[width=0.14\linewidth]{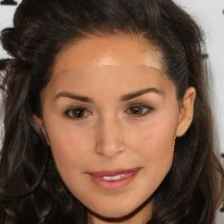} &
        \includegraphics[width=0.14\linewidth]{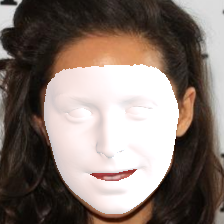} &
        \includegraphics[width=0.14\linewidth]{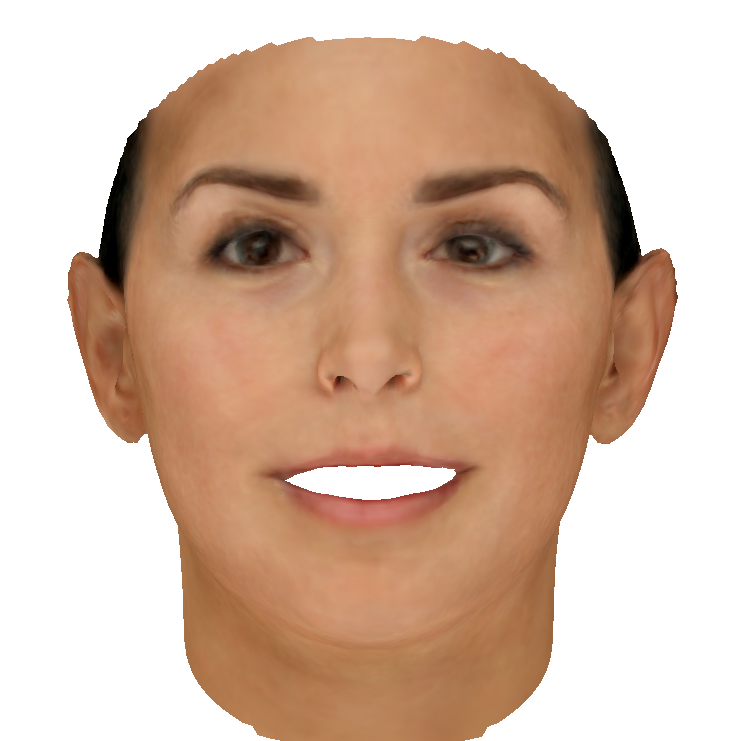} \\
    \end{tabular}

    \caption{Lighting and albedo results for MoFA  \cite{tewari2017mofa} and our 3DMM regression.}
    \label{fig:light}
\end{figure}

\subsubsection{Lighting and Albedo}

Fig. \ref{fig:light} shows the visualization of the albedo and lighting reconstruction results from MoFA and our 3DMM regression.
We set the meshes as white for a clear display.
Though the overall color looks similar between these two results, the lighting and albedo are different.
The colors in the overlay of MoFA  \cite{tewari2017mofa} are mostly from lighting, which leads to smooth and fair albedo.
In comparison, the albedo of our method is more faithful to the input faces.
%

\begin{figure}[t!]
    \centerline{\includegraphics[width=0.95\linewidth]{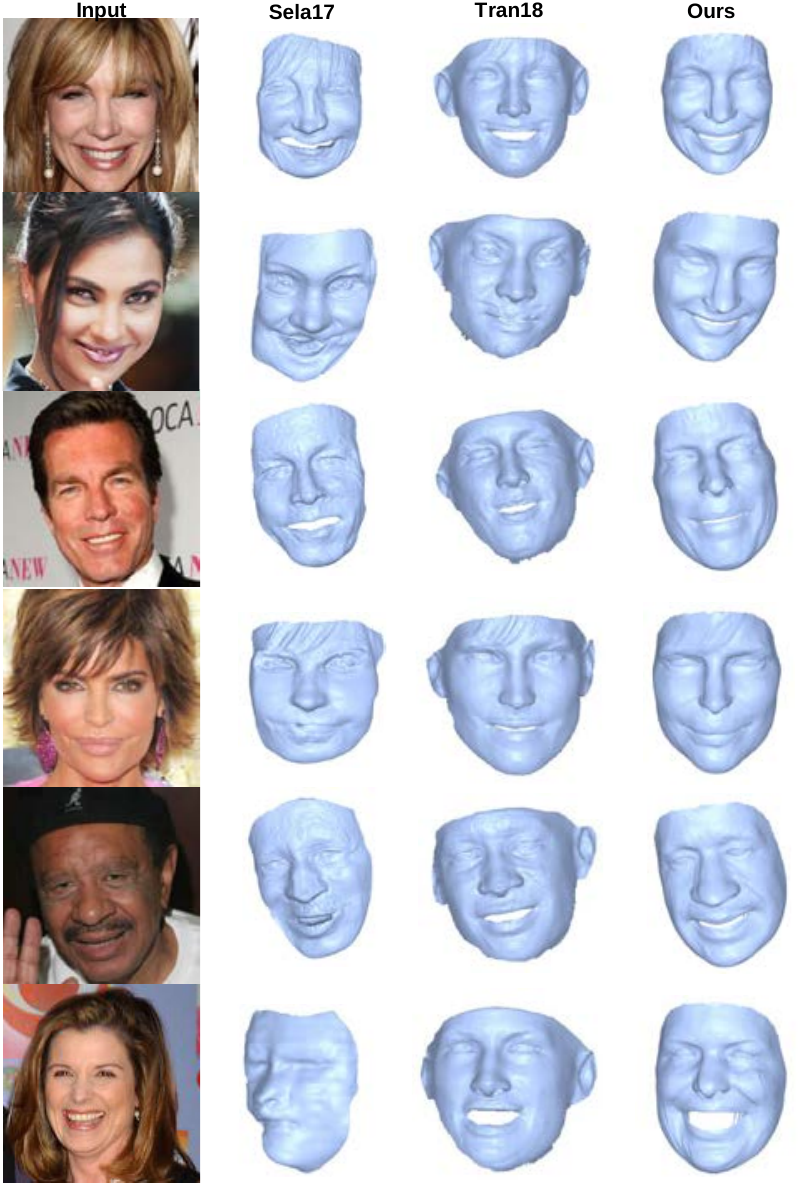}}
    \vspace{-3mm}
    \caption{Comparison to Sela17 \cite{sela2017unrestricted} and Tran18 \cite{tran2018extreme} of detailed reconstruction.}
    \label{fig:celeba}
\end{figure}

\begin{figure}[t!]
    \centering
    \includegraphics[width=1.0\linewidth]{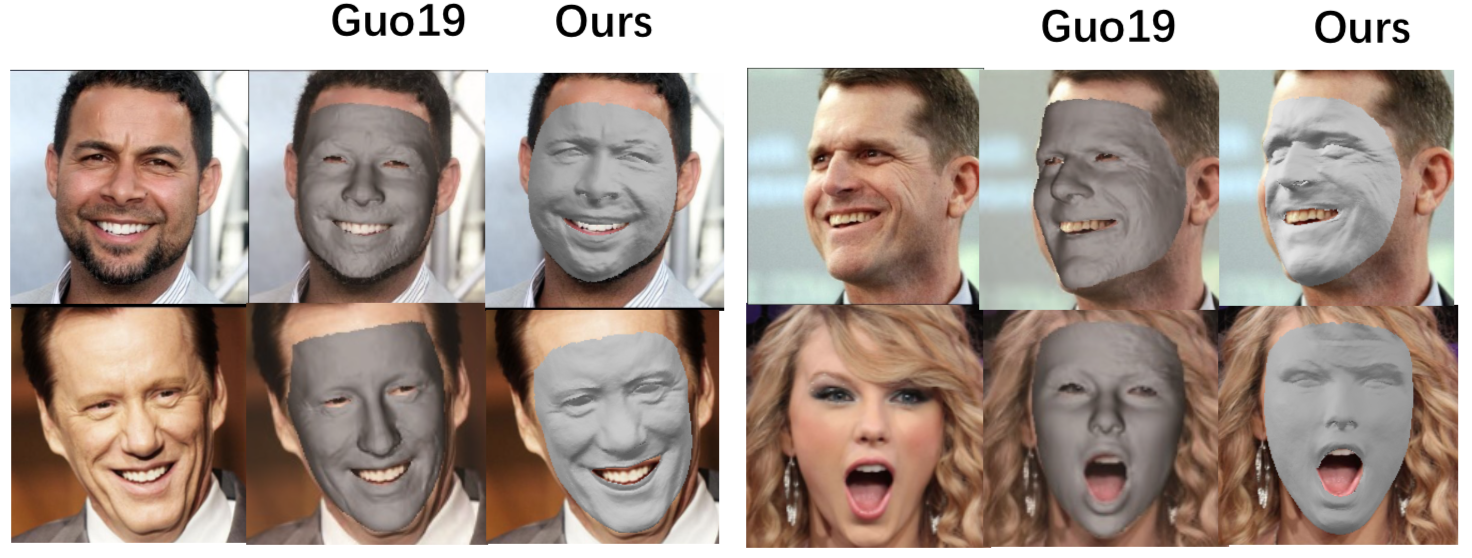}
    \caption{Comparison to Guo19 \cite{guo2019cnn} of detailed reconstruction.}
    \label{fig:comp_guo}
\end{figure}

\begin{figure}[t]
    \centering
    \footnotesize
    \setlength{\tabcolsep}{0.2em}
    \begin{tabular}{cccc}
         Close-up & Sela17 & Tran18 & Ours\\
        \includegraphics[width=0.2\linewidth]{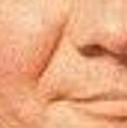} &
        \includegraphics[width=0.2\linewidth]{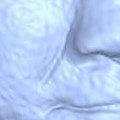} &
        \includegraphics[width=0.2\linewidth]{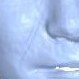} &
        \includegraphics[width=0.2\linewidth]{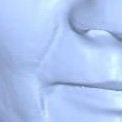} \\
        \includegraphics[width=0.2\linewidth]{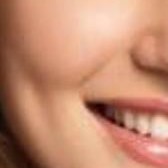} &
        \includegraphics[width=0.2\linewidth]{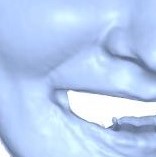} &
        \includegraphics[width=0.2\linewidth]{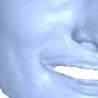} &
        \includegraphics[width=0.2\linewidth]{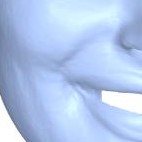} \\
        \includegraphics[width=0.2\linewidth]{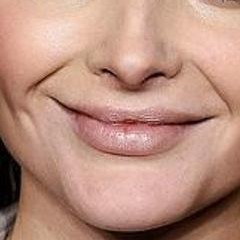} &
        \includegraphics[width=0.2\linewidth]{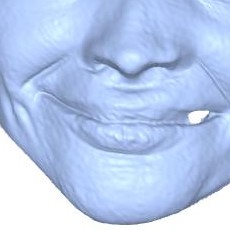} &
        \includegraphics[width=0.2\linewidth]{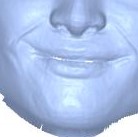} &
        \includegraphics[width=0.2\linewidth]{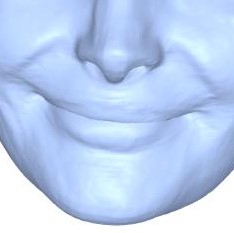} \\
    \end{tabular}

    \caption{Close-ups of detailed reconstruction results.}
    \label{fig:part}
\end{figure}

\begin{figure}[ht!bp]
\centering
    \includegraphics[width=0.95\linewidth]{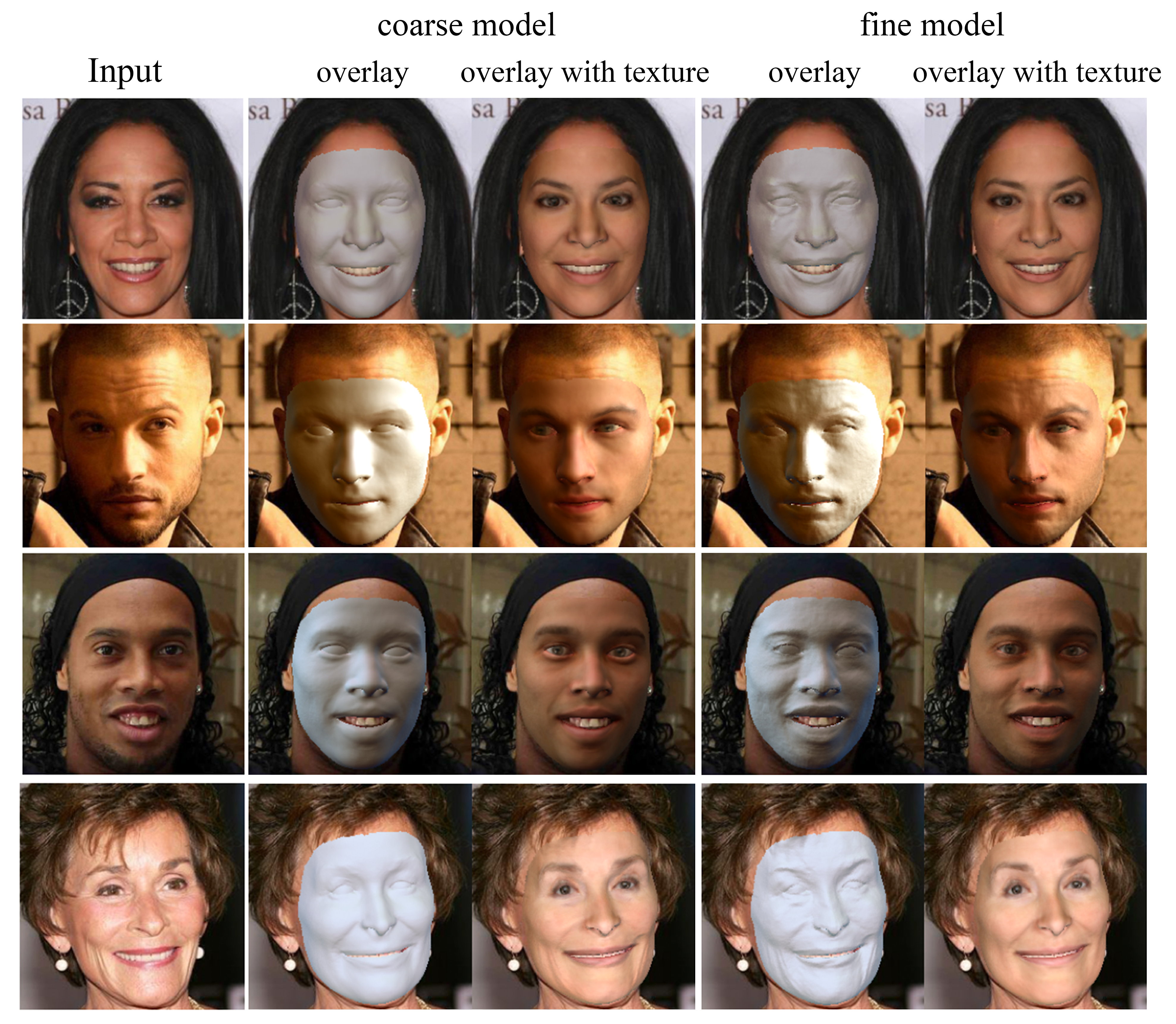}
    \caption{Visual comparison of the coarse model (3DMM regression) and fine model (detailed reconstruction), both obtained using our method.}
    \label{fig:abla}
\end{figure}

\subsection{Evaluation on Final Reconstruction}

We first show some examples in Fig. \ref{fig:abla} to demonstrate the differences between the coarse model obtained using our 3DMM regression and the fine model obtained using our full model. 
We can see that the fine model can retain more details and express more vivid facial characteristics than the 3DMM-based coarse model. 

We now show visual comparisons of our method to state-of-the-art methods \cite{sela2017unrestricted,tran2018extreme,guo2019cnn} for detailed 3D face reconstruction in Figs. \ref{fig:celeba} and \ref{fig:comp_guo}.
%
%
We observe that the 3D faces generated by Tran18 \cite{tran2018extreme} are often noisy, where high frequency information are spread all over the meshes regardless of the input images. For example, on the third row of Fig. \ref{fig:celeba}, the mouth region on the input face is smooth with salient texture, but the mouth of Tran18 is noisy. Meanwhile, the wrinkles on the cheeks are not reconstructed well according to the input image. Similar phenomena appear on other 3D face models which are not faithfully representing the input faces. The 3D models generated by Sela17 \cite{sela2017unrestricted} are sometimes distorted, where the results can not be regarded as faces. 
In comparison, our method effectively generates 3D models preserving the global shape and structure.  
Fig. \ref{fig:part} shows several additional examples of the close-ups. 
While Guo19 \cite{guo2019cnn} (Fig. \ref{fig:comp_guo}) has nice shape reconstruction results, our method reconstructs meshes with more details and are more faithful to input images.


We perform quantitative evaluation on the MICC Florence dataset  \cite{Bagdanov2011MICC} and the Face Recognition Grand Challenge V2 (FRGC2) dataset  \cite{phillips2005overview}. 
In MICC, we evaluate the shape reconstruction precision and in FGRC2 we evaluate the depth estimation errors.
We first evaluate the shape reconstruction precision on MICC dataset.
As we aim to model facial details, we select frontal video frames of each subject for evaluation.
The frontal frames only exist in the Indoor Cooperative condition.
We compute the point-to-point error between the reconstructed 3D faces and the ground truth scans.
We first crop the ground truth scan to 95mm around the tip of the nose.
Then, we run ICP algorithm with isotropic scale to find an alignment between the ground truth and the reconstruction before computing the point-to-point distances.

\begin{figure}[ht!]
    \centering
    \includegraphics[width=0.8\linewidth]{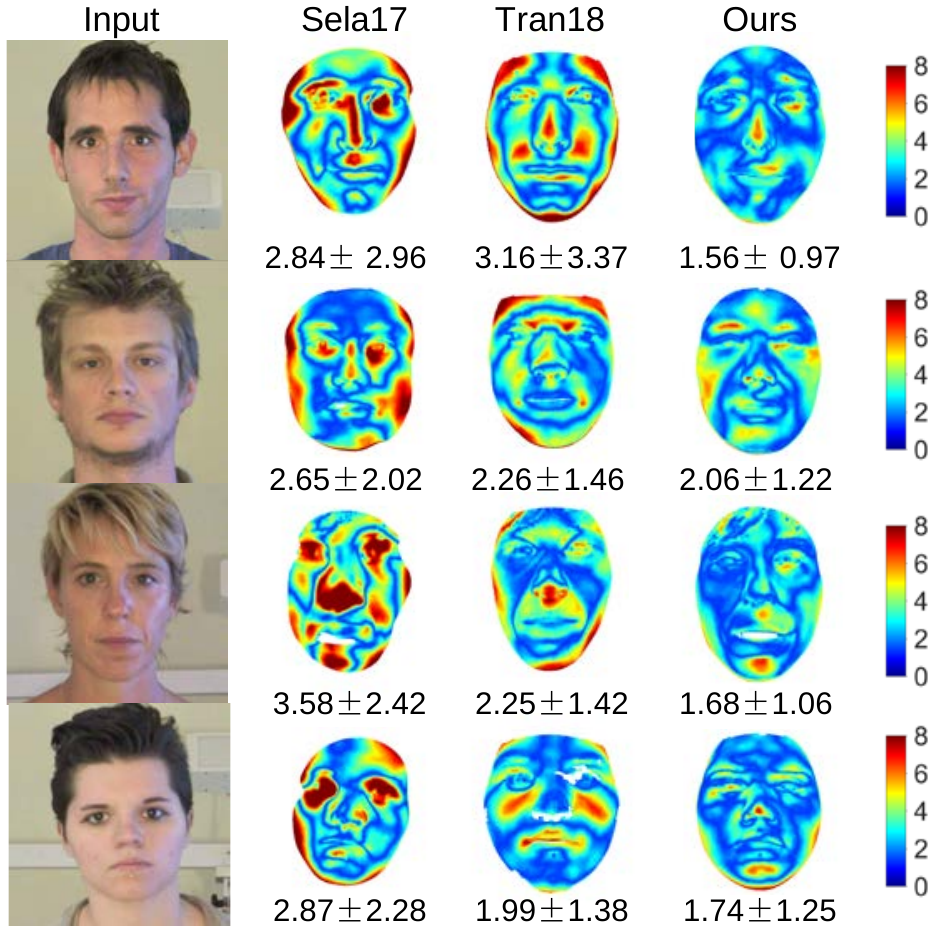}
    \vspace{-1mm}
    \caption{Examples of error maps for detailed reconstruction.}
    \label{fig:errormapmicc}
\end{figure}

\begin{figure}[ht!bp]
    \centering
    \includegraphics[width=0.9\linewidth]{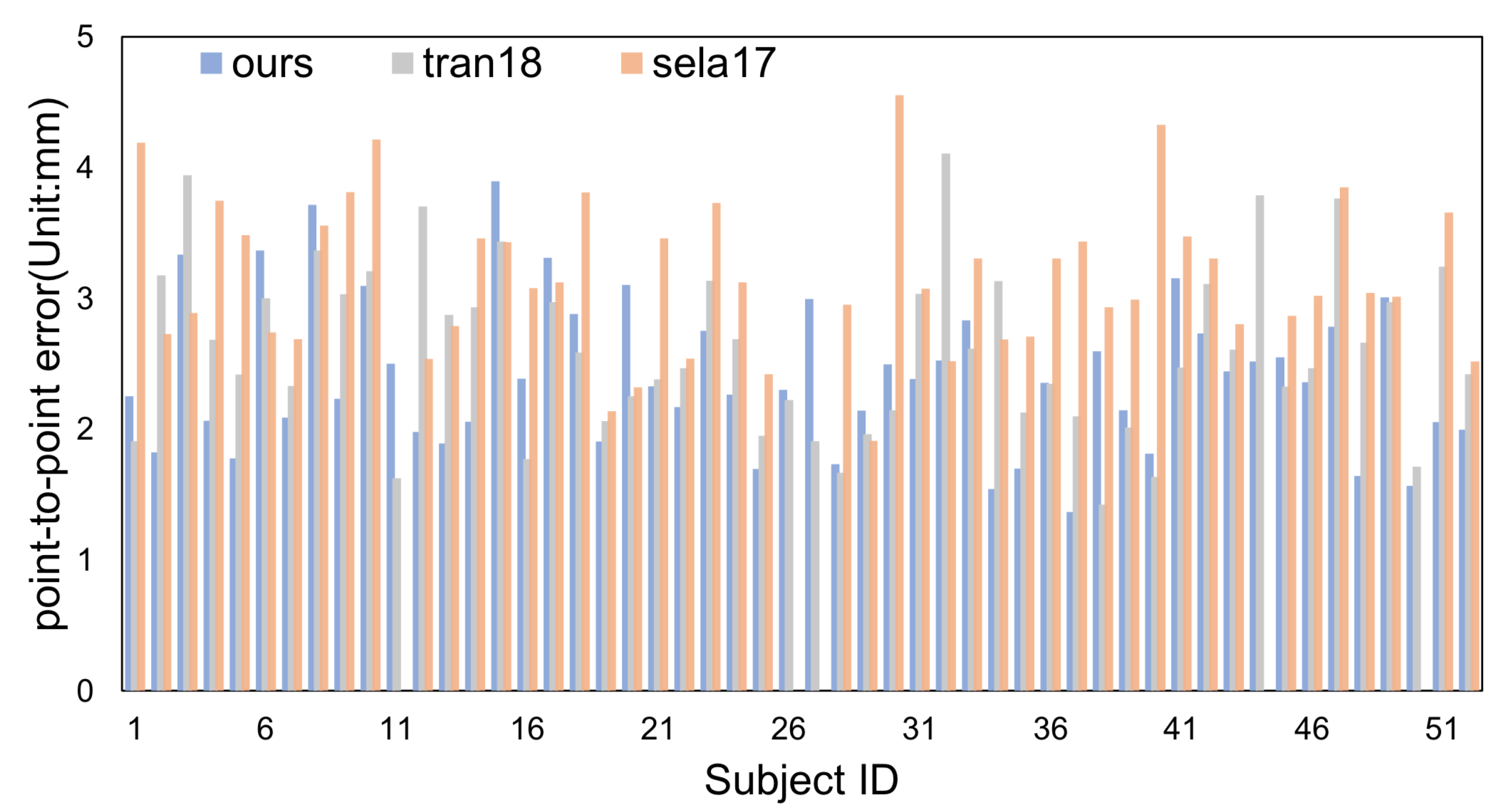}
    \vspace{-2mm}
    \caption{Barplots on MICC for different detailed reconstruction approaches.}
    \label{fig:barplotmicc}
\end{figure}

Table \ref{tab:micc2} shows the reconstruction error on MICC dataset compared with state-of-the-art methods \cite{sela2017unrestricted,tran2018extreme}.
The lower error demonstrates that our method can reconstruct fine details with higher accuracy and stability than the other methods.
During the evaluation process, Sela17  \cite{sela2017unrestricted} fails to reconstruct 5 subjects, while Tran18  \cite{tran2018extreme} and our method successfully reconstruct all subjects.
%
%
%
Fig. \ref{fig:errormapmicc} shows some error maps and Fig. \ref{fig:barplotmicc} shows the individual errors.
Sela17 is not robust compared with Tran18 and our method with higher error and standard deviation metrics.
On the contrary, our method stably produces state-of-the-art results.

\begin{figure}[ht!bp]
    \vspace{-2mm}
    \centering
    \includegraphics[width=0.9\linewidth]{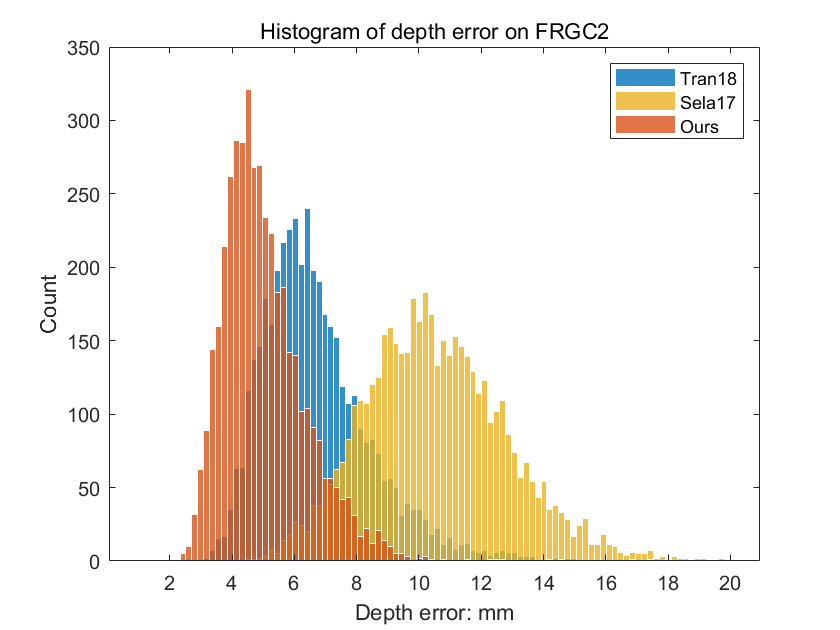}
    \vspace{-3mm}
    \caption{Depth error histogram on FRGC2 \cite{phillips2005overview} for different approaches.}
    \label{fig:hist_depth}
\end{figure}

Besides MICC, we also evaluate on the FRGC2 datasets where the depth of the input face images are estimated.
The Face Recognition Grand Challenge V2 (FRGC2) dataset  \cite{phillips2005overview} includes 4,950 high resolution images of 688 identities with corresponding depth maps. We evaluate the depth estimation results generated by all the methods on the FRGC2.
To calculate the depth error, we first scale the depth estimation of each method to fit the ground truth depths in min-max ranges.
Then the mean distance between the two depth maps at valid pixel positions provided by a fixed binary mask are computed as depth error.
Table \ref{tab:frgc} shows the depth estimation results for these methods, while Fig. \ref{fig:hist_depth} displays the corresponding histogram.
our method achieve lowest mean and standard deviation in depth errors compared with the other approaches.
The low depth errors indicate that the proposed model generate 3D faces with higher accuracy.

\begin{figure}[ht!]
    \centering
    \includegraphics[width=0.7\linewidth]{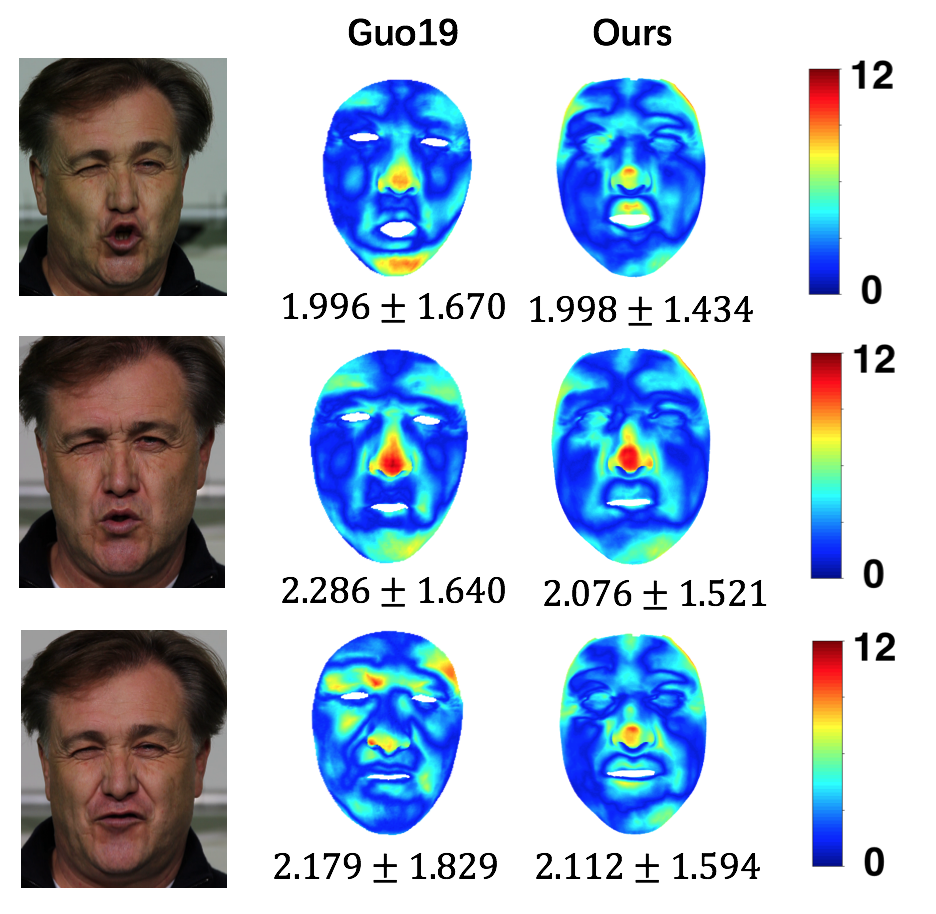}
    \vspace{-2mm}
    \caption{Examples of error maps for detailed reconstruction.}
    \label{fig:errormapfacecap}
\end{figure}

Fig. \ref{fig:errormapfacecap} shows several error map examples of our method and Guo19 \cite{guo2019cnn} on FaceCap \cite{facecap13}. The dataset consists of stereo captures and the corresponding 3D reconstructions, which are used as ground truth data. our method achieves comparable results with Guo19 in 3D face detail reconstruction, though both approaches have large error on noses.

\begin{table}[t!]
\begin{center}
\begin{tabular}{c|c|c|c}
\hline
\textbf{Method} & \textbf{Sela17 \cite{sela2017unrestricted}} & \textbf{Tran18 \cite{tran2018extreme}}  & \textbf{Ours} \\
\hline\hline
Error(mm)  & $3.19\pm0.58$ & $2.61\pm0.66$ & $\mathbf{2.41\pm0.62}$\\
\hline
\end{tabular}
\end{center}
\caption{Point-to-point error on MICC Florence \cite{Bagdanov2011MICC} for different detailed reconstruction approaches.}
\label{tab:micc2}
\end{table}

\begin{table}[t]
\begin{center}
\begin{tabular}{c|c|c|c}
\hline
\textbf{Method} & \textbf{Sela17 \cite{sela2017unrestricted}} & \textbf{Tran18 \cite{tran2018extreme}} & \textbf{Ours} \\
\hline\hline
Error(mm) & $10.55\pm2.34$  &  $6.73\pm1.85$   & $\mathbf{5.05\pm1.44}$\\
\hline
\end{tabular}
\end{center}
\caption{Depth error on FRGC2 \cite{phillips2005overview} for different detailed reconstruction approaches.}
\label{tab:frgc}
\end{table}

\begin{figure}[t]
    \vspace{-1mm}
    \includegraphics[width=\linewidth]{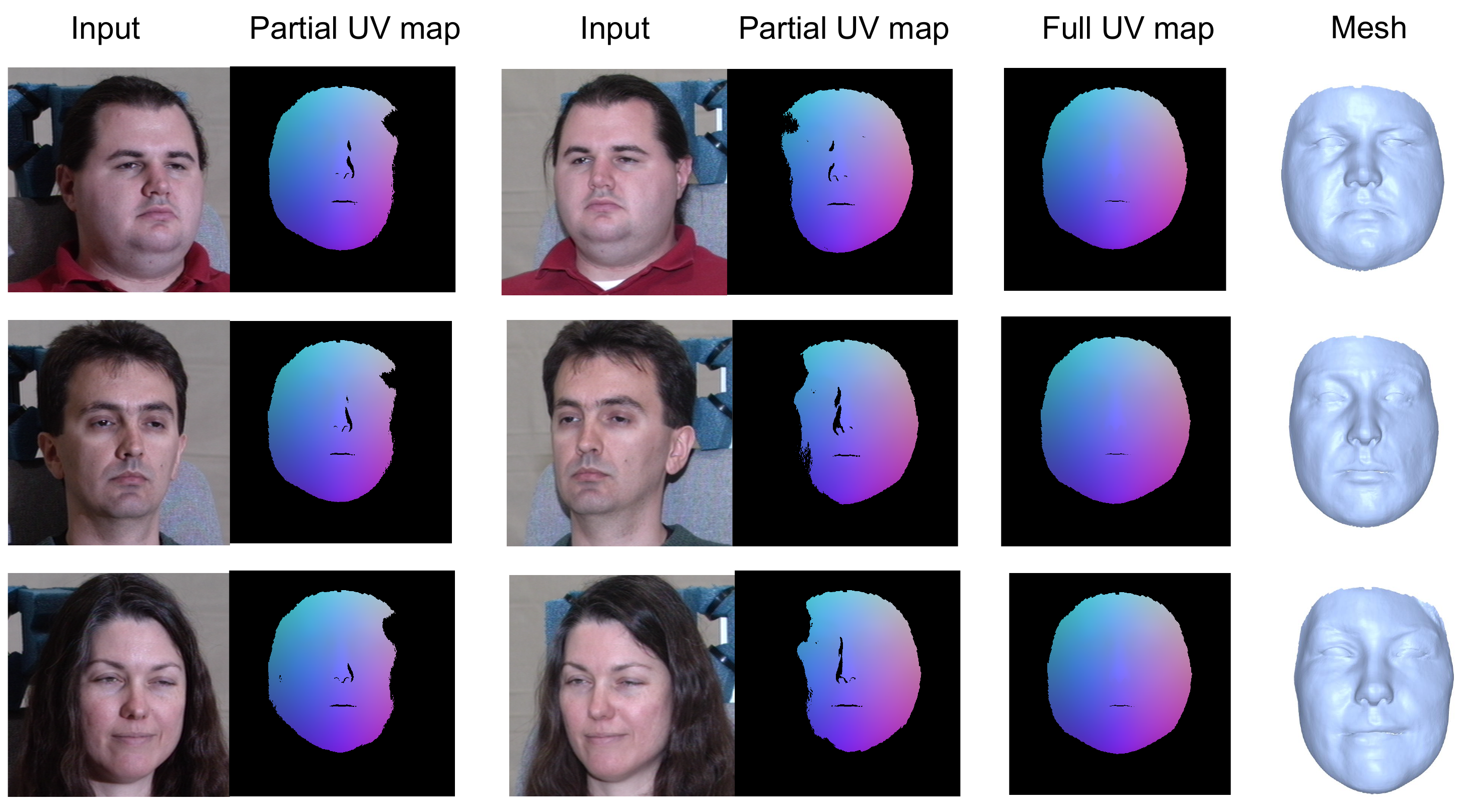}
    \vspace{-5mm}
    \caption{Though our method is trained with single-view images, it can be adopted for multi-view reconstruction by merging two partial UV position maps into a full UV position map.}
    \label{fig:blend}
\end{figure}

\subsection{Application}

An additional advantage of employing UV position maps  \cite{feng2018joint} for reconstruction is that it enables easier integration of 3D reconstructions from different views of a same face. The UV maps from different views can be easily combined by a simple blending in UV-space.
Thus the combined full UV map can represent a complete 3D face model that are visible in different views.
The induced detailed 3D reconstruction are more complete compared to depth map based representations.
Fig. \ref{fig:blend} shows several examples of blending two partial UV maps from two views of a same face.
In practice, to handle difficult lighting conditions or blurry faces, some preprocessing approach can be used \cite{song-ijcv19-jfh}.


\begin{figure}[t]
    \centering
    \footnotesize
    \setlength{\tabcolsep}{0.1em}
    \begin{tabular}{cccccc}
         \multicolumn{2}{c}{\textbf{Extreme expression}}& \multicolumn{2}{c}{\textbf{Occlusion}} & \multicolumn{2}{c}{\textbf{Large pose}} \\
         Input & Shape & Input & Shape & Input & Shape \\
        \includegraphics[width=0.16\linewidth]{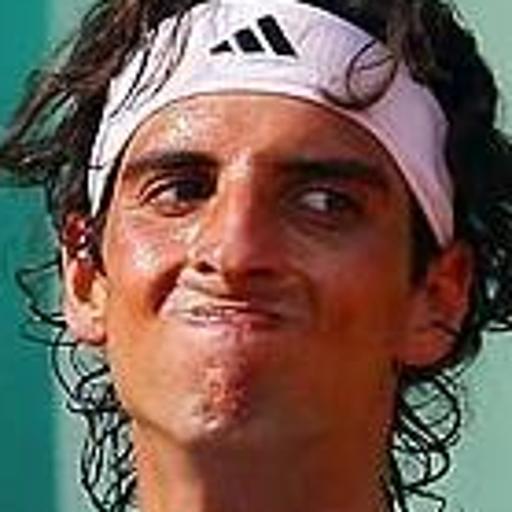} &
        \includegraphics[width=0.16\linewidth]{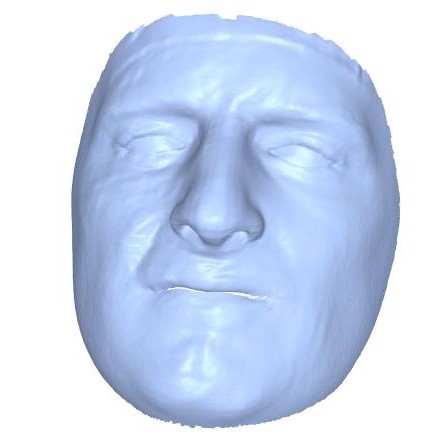} &
        \includegraphics[width=0.16\linewidth]{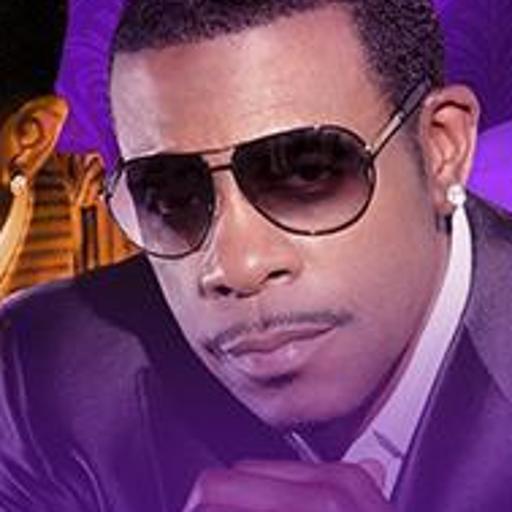} &
        \includegraphics[width=0.16\linewidth]{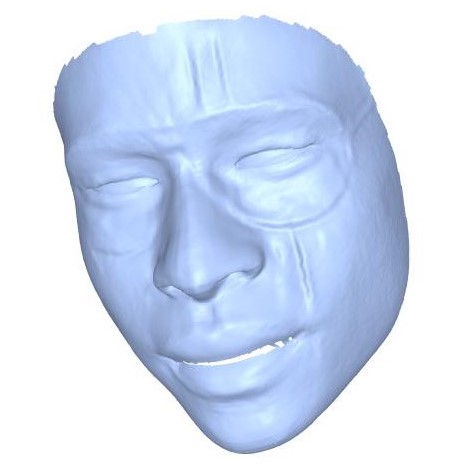} &
        \includegraphics[width=0.16\linewidth]{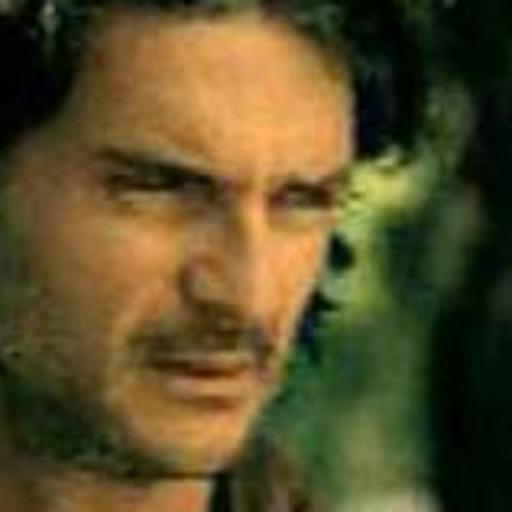} &
        \includegraphics[width=0.16\linewidth]{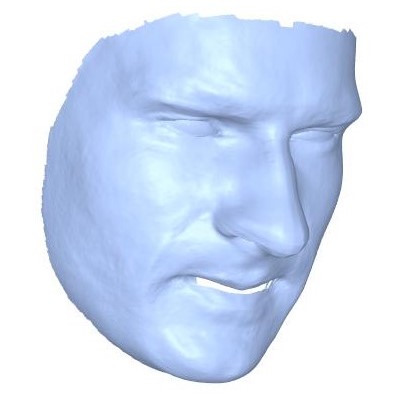}\\
        \includegraphics[width=0.16\linewidth]{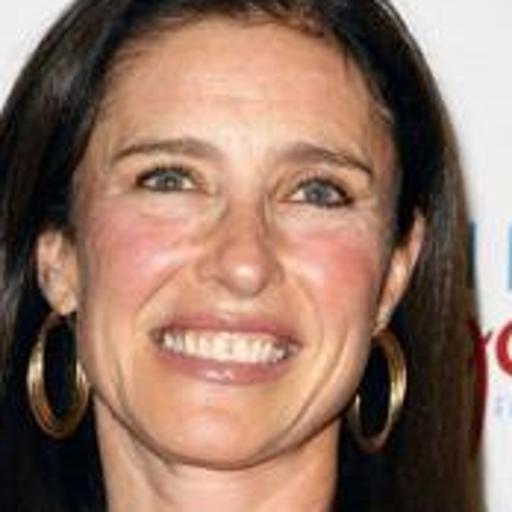} &
        \includegraphics[width=0.16\linewidth]{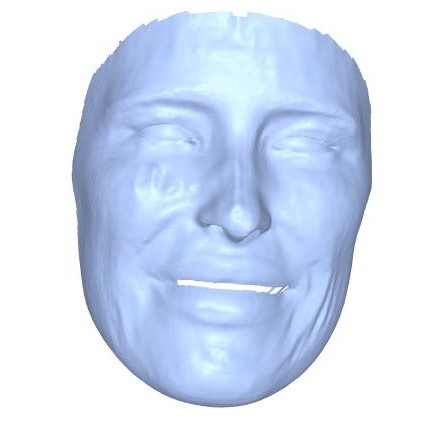} &
        \includegraphics[width=0.16\linewidth]{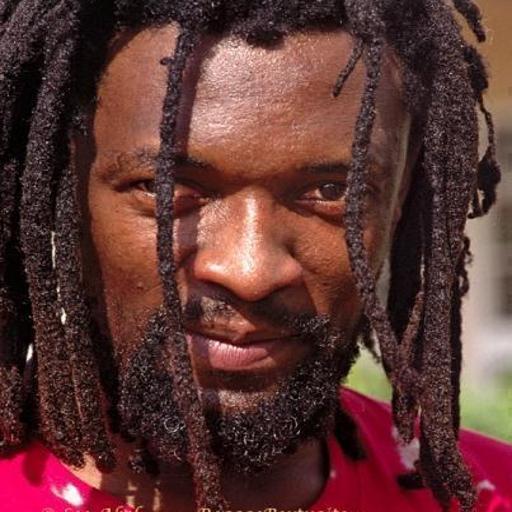} &
        \includegraphics[width=0.16\linewidth]{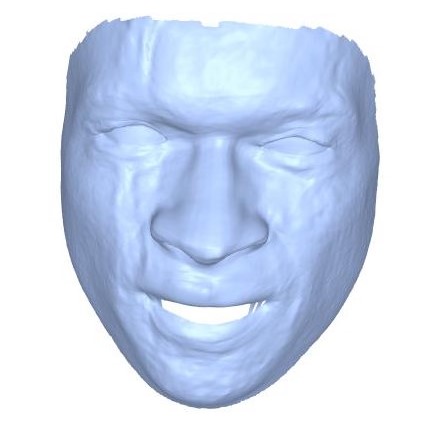} &
        \includegraphics[width=0.16\linewidth]{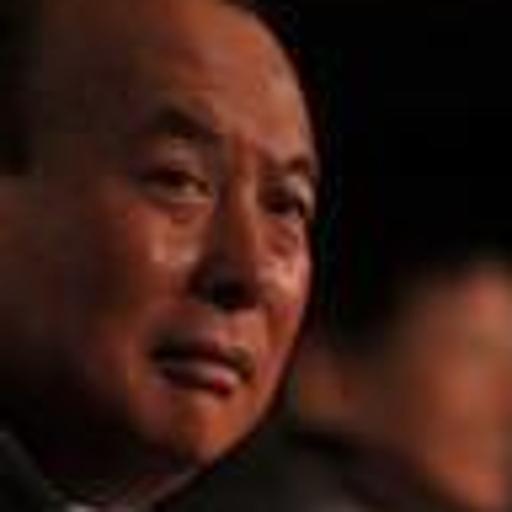} &
        \includegraphics[width=0.16\linewidth]{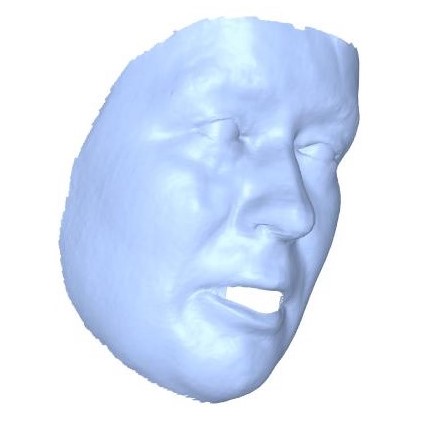}\\
        \includegraphics[width=0.16\linewidth]{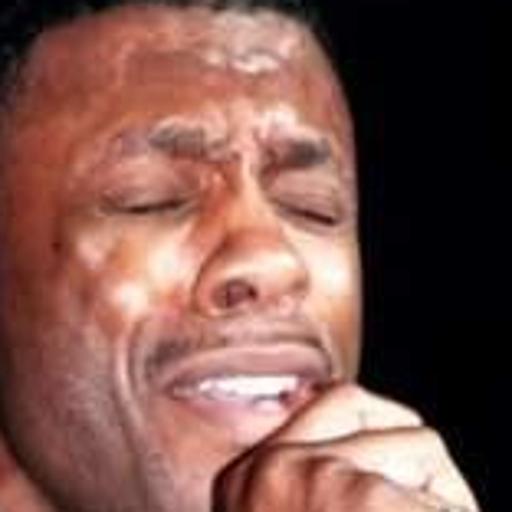} &
        \includegraphics[width=0.16\linewidth]{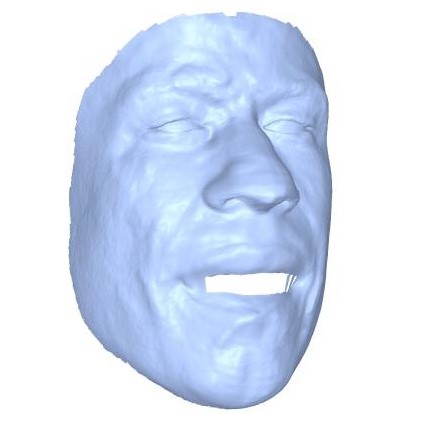} &
        \includegraphics[width=0.16\linewidth]{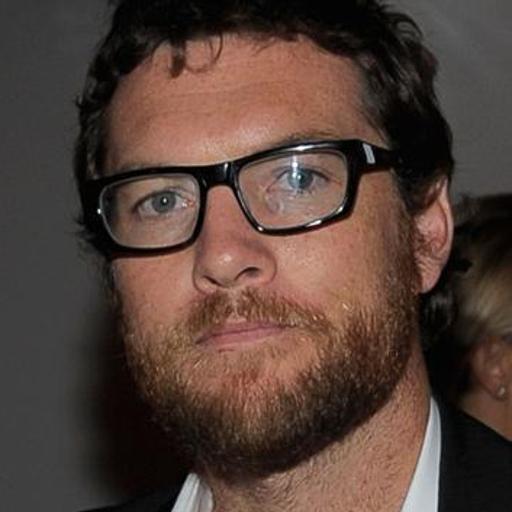} &
        \includegraphics[width=0.16\linewidth]{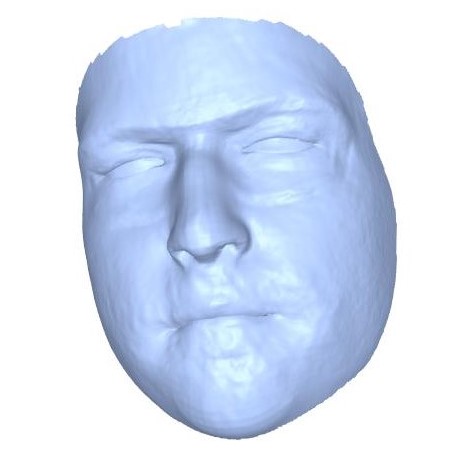} &
        \includegraphics[width=0.16\linewidth]{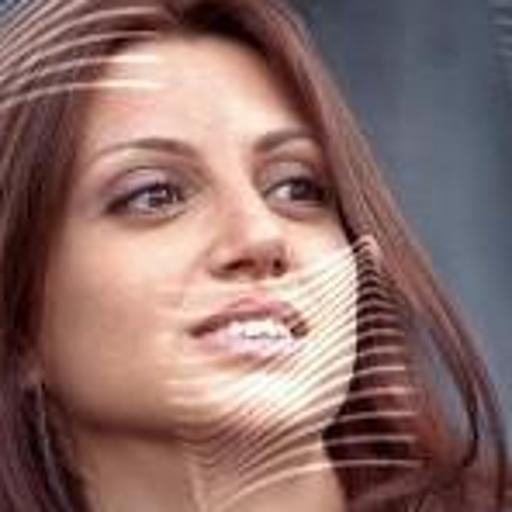} &
        \includegraphics[width=0.16\linewidth]{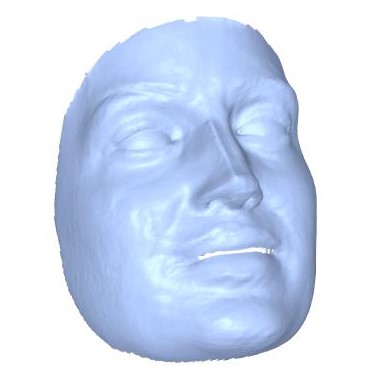}\\
    \end{tabular}

    \caption{Limitations of our method. The proposed model have troubles dealing with partial occlusion, extreme poses and expressions.}
    \label{fig:limit}
\end{figure}

\subsection{Limitation Analysis}

Fig. \ref{fig:limit} shows some failure cases of the proposed model under scenarios including large expression, occlusion and large pose. Since the proposed model are not trained with specific policy to deal with these conditions, the reconstruction quality can not be guaranteed if these situation occur.

\section{Conclusion}

We propose a detailed 3D face reconstruction framework with self-supervised learning. We first use a coarse 3DMM encoder to regress 3DMM parameters. When learning the 3DMM encoder, we incorporate measurements of multiple loss terms ranging from the pixel-wise similarity to the global facial perception. After learning the 3DMM parameters, we unwrap both input face image and 3DMM texture into UV space where all the faces are precisely aligned. The details from the inputs are effectively transferred to the 3D model as the aligned facial details facilitate the learning process. Experiments on various datasets demonstrate our method performs favorably against state-of-the-art 3D face modeling approaches.

\bibliographystyle{IEEEtran}
\bibliography{ref}

\begin{thebibliography}{10}
\providecommand{\url}[1]{#1}
\csname url@samestyle\endcsname
\providecommand{\newblock}{\relax}
\providecommand{\bibinfo}[2]{#2}
\providecommand{\BIBentrySTDinterwordspacing}{\spaceskip=0pt\relax}
\providecommand{\BIBentryALTinterwordstretchfactor}{4}
\providecommand{\BIBentryALTinterwordspacing}{\spaceskip=\fontdimen2\font plus
\BIBentryALTinterwordstretchfactor\fontdimen3\font minus
  \fontdimen4\font\relax}
\providecommand{\BIBforeignlanguage}[2]{{%
\expandafter\ifx\csname l@#1\endcsname\relax
\typeout{** WARNING: IEEEtran.bst: No hyphenation pattern has been}%
\typeout{** loaded for the language `#1'. Using the pattern for}%
\typeout{** the default language instead.}%
\else
\language=\csname l@#1\endcsname
\fi
#2}}
\providecommand{\BIBdecl}{\relax}
\BIBdecl

\bibitem{blanz1999morphable}
V.~Blanz and T.~Vetter, ``A morphable model for the synthesis of 3d faces,'' in
  \emph{ACM Transactions on Graphics (Proceedings of SIGGRAPH)}, 1999.

\bibitem{blanz2003face}
------, ``Face recognition based on fitting a 3d morphable model,'' \emph{IEEE
  Transactions on Pattern Analysis and Machine Intelligence}, 2003.

\bibitem{romdhani2005estimating}
S.~Romdhani and T.~Vetter, ``Estimating 3d shape and texture using pixel
  intensity, edges, specular highlights, texture constraints and a prior,'' in
  \emph{IEEE Conference on Computer Vision and Pattern Recognition}, 2005.

\bibitem{zhu2016face}
X.~Zhu, Z.~Lei, X.~Liu, H.~Shi, and S.~Z. Li, ``Face alignment across large
  poses: A 3d solution,'' in \emph{IEEE Conference on Computer Vision and
  Pattern Recognition}, 2016.

\bibitem{tran2017regressing}
A.~T. Tran, T.~Hassner, I.~Masi, and G.~Medioni, ``Regressing robust and
  discriminative 3d morphable models with a very deep neural network,'' in
  \emph{IEEE Conference on Computer Vision and Pattern Recognition}, 2017.

\bibitem{tewari2017mofa}
A.~Tewari, M.~Zollh{\"o}fer, H.~Kim, P.~Garrido, F.~Bernard, P.~P{\'e}rez, and
  C.~Theobalt, ``Mofa: Model-based deep convolutional face autoencoder for
  unsupervised monocular reconstruction,'' in \emph{IEEE International
  Conference on Computer Vision}, 2017.

\bibitem{kim2018inversefacenet}
H.~Kim, M.~Zollh{\"o}fer, A.~Tewari, J.~Thies, C.~Richardt, and C.~Theobalt,
  ``Inversefacenet: Deep monocular inverse face rendering,'' in \emph{IEEE
  Conference on Computer Vision and Pattern Recognition}, 2018.

\bibitem{genova2018unsupervised}
K.~Genova, F.~Cole, A.~Maschinot, A.~Sarna, D.~Vlasic, and W.~T. Freeman,
  ``Unsupervised training for 3d morphable model regression,'' in \emph{IEEE
  Conference on Computer Vision and Pattern Recognition}, 2018.

\bibitem{Tewari_2018_CVPR}
A.~Tewari, M.~Zollhöfer, P.~Garrido, F.~Bernard, H.~Kim, P.~Pérez, and
  C.~Theobalt, ``Self-supervised multi-level face model learning for monocular
  reconstruction at over 250 hz,'' in \emph{IEEE Conference on Computer Vision
  and Pattern Recognition}, 2018.

\bibitem{tran2018extreme}
A.~T. Tran, T.~Hassner, I.~Masi, E.~Paz, Y.~Nirkin, and G.~Medioni, ``Extreme
  {3D} face reconstruction: Seeing through occlusions,'' in \emph{IEEE
  Conference on Computer Vision and Pattern Recognition}, 2018.

\bibitem{richardson20163d}
E.~Richardson, M.~Sela, and R.~Kimmel, ``3d face reconstruction by learning
  from synthetic data,'' in \emph{International Conference on 3D Vision}, 2016.

\bibitem{dou2017end}
P.~Dou, S.~K. Shah, and I.~A. Kakadiaris, ``End-to-end 3d face reconstruction
  with deep neural networks,'' in \emph{IEEE Conference on Computer Vision and
  Pattern Recognition}, 2017.

\bibitem{jackson2017large}
A.~S. Jackson, A.~Bulat, V.~Argyriou, and G.~Tzimiropoulos, ``Large pose 3d
  face reconstruction from a single image via direct volumetric cnn
  regression,'' in \emph{IEEE International Conference on Computer Vision},
  2017.

\bibitem{sela2017unrestricted}
M.~Sela, E.~Richardson, and R.~Kimmel, ``Unrestricted facial geometry
  reconstruction using image-to-image translation,'' in \emph{IEEE
  International Conference on Computer Vision}, 2017.

\bibitem{richardson2017learning}
E.~Richardson, M.~Sela, R.~Or-El, and R.~Kimmel, ``Learning detailed face
  reconstruction from a single image,'' in \emph{IEEE Conference on Computer
  Vision and Pattern Recognition}, 2017.

\bibitem{zollhofer2018state}
M.~Zollh{\"o}fer, J.~Thies, P.~Garrido, D.~Bradley, T.~Beeler, P.~P{\'e}rez,
  M.~Stamminger, M.~Nie{\ss}ner, and C.~Theobalt, ``State of the art on
  monocular 3d face reconstruction, tracking, and applications,'' in
  \emph{Computer Graphics Forum}, 2018.

\bibitem{cao2014facewarehouse}
C.~Cao, Y.~Weng, S.~Zhou, Y.~Tong, and K.~Zhou, ``Facewarehouse: A 3d facial
  expression database for visual computing,'' \emph{IEEE Transactions on
  Visualization and Computer Graphics}, 2014.

\bibitem{garrido2013reconstructing}
P.~Garrido, L.~Valgaerts, C.~Wu, and C.~Theobalt, ``Reconstructing detailed
  dynamic face geometry from monocular video.'' in \emph{ACM Transactions on
  Graphics (Proceedings of SIGGRAPH)}, 2013.

\bibitem{thies2016face2face}
J.~Thies, M.~Zollhofer, M.~Stamminger, C.~Theobalt, and M.~Nie{\ss}ner,
  ``Face2face: Real-time face capture and reenactment of rgb videos,'' in
  \emph{IEEE Conference on Computer Vision and Pattern Recognition}, 2016.

\bibitem{wu2019mvf}
F.~Wu, L.~Bao, Y.~Chen, Y.~Ling, Y.~Song, S.~Li, K.~N. Ngan, and W.~Liu,
  ``Mvf-net: Multi-view 3d face morphable model regression,'' in \emph{IEEE
  Conference on Computer Vision and Pattern Recognition}, 2019.

\bibitem{guo2019cnn}
Y.~{Guo}, j.~{zhang}, J.~{Cai}, B.~{Jiang}, and J.~{Zheng}, ``Cnn-based
  real-time dense face reconstruction with inverse-rendered photo-realistic
  face images,'' \emph{IEEE Transactions on Pattern Analysis and Machine
  Intelligence}, 2019.

\bibitem{Parkhi15vggface}
O.~M. Parkhi, A.~Vedaldi, and A.~Zisserman, ``Deep face recognition,'' in
  \emph{British Machine Vision Conference}, 2015.

\bibitem{bulat2017far}
A.~Bulat and G.~Tzimiropoulos, ``How far are we from solving the 2d \& 3d face
  alignment problem?(and a dataset of 230,000 3d facial landmarks),'' in
  \emph{IEEE International Conference on Computer Vision}, 2017.

\bibitem{liu2015celeba}
Z.~Liu, P.~Luo, X.~Wang, and X.~Tang, ``Deep learning face attributes in the
  wild,'' in \emph{IEEE International Conference on Computer Vision}, 2015.

\bibitem{LFWTech}
G.~B. Huang, M.~Ramesh, T.~Berg, and E.~Learned-Miller, ``Labeled faces in the
  wild: A database for studying face recognition in unconstrained
  environments,'' Tech. Rep., 2007.

\bibitem{isola2017image}
P.~Isola, J.-Y. Zhu, T.~Zhou, and A.~A. Efros, ``Image-to-image translation
  with conditional adversarial networks,'' in \emph{IEEE conference on computer
  vision and pattern recognition}, 2017, pp. 1125--1134.

\bibitem{Bagdanov2011MICC}
A.~D. Bagdanov, A.~Del~Bimbo, and I.~Masi, ``The florence 2d/3d hybrid face
  dataset,'' in \emph{Joint ACM Workshop on Human Gesture and Behavior
  Understanding}, 2011.

\bibitem{phillips2005overview}
P.~J. Phillips, P.~J. Flynn, T.~Scruggs, K.~W. Bowyer, J.~Chang, K.~Hoffman,
  J.~Marques, J.~Min, and W.~Worek, ``Overview of the face recognition grand
  challenge,'' in \emph{IEEE Conference on Computer Vision and Pattern
  Recognition}, 2005.

\bibitem{facecap13}
P.~Garrido, L.~Valgaerts, C.~Wu, and C.~Theobalt, ``Reconstructing detailed
  dynamic face geometry from monocular video,'' in \emph{ACM Transactions on
  Graphics (Proceedings of SIGGRAPH)}, 2013.

\bibitem{feng2018joint}
Y.~Feng, F.~Wu, X.~Shao, Y.~Wang, and X.~Zhou, ``Joint 3d face reconstruction
  and dense alignment with position map regression network,'' in \emph{European
  Conference on Computer Vision}, 2018.

\bibitem{song-ijcv19-jfh}
Y.~Song, J.~Zhang, L.~Gong, S.~He, L.~Bao, J.~Pan, Q.~Yang, and M.-H. Yang,
  ``Joint face hallucination and deblurring via structure generation and detail
  enhancement,'' \emph{International Journal of Computer Vision}, 2019.

\end{thebibliography}


\end{document}